%% file: paper.tex
\documentclass{article}

\usepackage[utf8]{inputenc} %
\usepackage[T1]{fontenc}    %
\usepackage[table]{xcolor}  %
\definecolor{forest}  {rgb}{0,.4,0} 
\definecolor{midnight}  {rgb}{0,0,.5} 
\usepackage[pdftex,plainpages=false,pdfpagelabels,colorlinks=true,urlcolor=midnight,citecolor=forest]{hyperref}
\usepackage{graphicx}
\usepackage{url}            %
\usepackage{booktabs}       %
\usepackage{amsfonts}       %
\usepackage{nicefrac}       %
\usepackage{microtype}      %
\usepackage{lipsum}		%
\usepackage{natbib}
\usepackage{doi}

\usepackage{multirow}
\usepackage[utf8]{inputenc} %
\usepackage[T1]{fontenc}    %
\usepackage{hyperref}       %
\usepackage{url}            %
\usepackage{booktabs}       %
\usepackage{amsfonts}       %
\usepackage{nicefrac}       %
\usepackage{microtype}      %
\usepackage{xcolor}         %
\usepackage{caption}
\usepackage{subcaption}
\usepackage{mathtools}
\newcommand{\C}{\mathbb{C}}
\newcommand{\R}{\mathbb{R}}
\DeclareMathOperator*{\argmin}{arg\,min}
\usepackage[capitalize,noabbrev]{cleveref}
\usepackage[linesnumbered,boxed,algoruled, noline]{algorithm2e}
\usepackage{multirow}
\usepackage{pifont}%
\newcommand{\cmark}{\ding{51}}%
\newcommand{\xmark}{\ding{55}}%

\usepackage[capitalize,noabbrev]{cleveref}
\usepackage[letterpaper,top=2cm,bottom=2cm,left=3cm,right=3cm,marginparwidth=1.75cm]{geometry}

\title{Rotation-Invariant Random Features Provide a Strong Baseline for Machine Learning on 3D Point Clouds}

\usepackage{authblk}

\author[1]{Owen Melia \thanks{meliao@uchicago.edu}}
\author[1]{Eric Jonas \thanks{ericj@uchicago.edu}}
\author[1,2]{Rebecca Willett \thanks{willett@uchicago.edu}}
\affil[1]{Department of Computer Science, University of Chicago, US}
\affil[2]{Department of Statistics, University of Chicago, US}
\date{}                   
\begin{document}
\maketitle

\begin{abstract}
\input{abstract.tex}
\end{abstract}

\input{introduction.tex}

\input{related_work.tex}
\input{rotational_invariance.tex}
\input{our_method.tex}
\input{experiments.tex}
\input{conclusion.tex}
\input{broader_impacts.tex}
\input{acknowledgements.tex}

\bibliographystyle{unsrtnat}
\bibliography{IRF_citations_2022-11-07}  

\appendix
\input{appendix_integration_details.tex}

\input{appendix_integration_details.tex}
\input{appendix_connection_with_ACE.tex}
\input{appendix_representation_theory.tex}
\input{appendix_extra_latency_experiments.tex}

\input{appendix_extra_figures.tex}
\input{appendix_solving_least_squares_problems.tex}

\end{document}

%% file: abstract.tex
Rotational invariance is a popular inductive bias used by many fields in machine learning, such as computer vision and machine learning for quantum chemistry. 
Rotation-invariant machine learning methods set the state of the art for many tasks, including molecular property prediction and 3D shape classification. 
These methods generally either rely on task-specific rotation-invariant features, or they use general-purpose deep neural networks which are complicated to design and train.
However, it is unclear whether the success of these methods is primarily due to the rotation invariance or the deep neural networks. 
To address this question, we suggest a simple and general-purpose method for learning rotation-invariant functions of three-dimensional point cloud data 
using a random features approach. Specifically, we extend the random features method of \citet{rahimi_random_2007} by deriving a version that is invariant to three-dimensional rotations and showing that it is fast to evaluate on point cloud data. 
We show through experiments that our method matches or outperforms the performance of general-purpose rotation-invariant neural networks on standard molecular property prediction benchmark datasets QM7 and QM9. 
We also show that our method is general-purpose and provides a rotation-invariant baseline on the ModelNet40 shape classification task. 
Finally, we show that our method has an order of magnitude smaller prediction latency than competing kernel methods.

%% file: introduction.tex
\section{Introduction}

Many common prediction tasks where the inputs are three-dimensional physical objects are known to be rotation-invariant; the ground-truth label does not change when the object is rotated. 
Building rotation invariance into machine learning models is an important inductive bias for such problems. 
The common intuition is that restricting the learning process to rotation-invariant models will remove any possibility of poor generalization performance due to rotations of test samples and may improve sample efficiency by reducing the effective complexity of learned models. 
These ideas have inspired a line of research begun by \citet{kondor_clebsch-gordan_2018,cohen_spherical_2018,esteves_learning_2018} into building general-purpose deep neural network architectures that are invariant to rotations of their input. 
However, in these studies, it is not clear whether the reported high accuracies are due to the expressive power of the neural networks or primarily attributable to rotation invariance. 
We introduce a rotation-invariant random feature model which helps us explore the impact of rotation invariance alone, outside of the neural network framework. Our method is general-purpose and does not require expert knowledge for feature or architecture design. For certain prediction tasks, our proposed method can be computed with very low prediction latency with only a small reduction in accuracy compared to neural network methods, making it a viable method in a range of applications.

In this paper, we consider prediction problems that are \textit{rotation-invariant}, that is, the ground-truth response $f^*(x) =y$ does not change when an arbitrary rotation is applied to the input $x$.
We represent the input data $x$ as a 3D point cloud, an unordered set of points in $\R^3$ possibly with accompanying labels. Common examples of rotation-invariant prediction problems on 3D point clouds include molecular property prediction, where the point cloud consists of the positions of a molecule's constituent atoms, and 3D shape classification, where the points are sampled from the surface of an object. \cref{sec:related_work} provides a detailed overview of methods used for rotation-invariant prediction, their underlying insights, and  their various advantages and disadvantages.

\begin{figure}
    \centering
    \includegraphics[width=\textwidth, trim={0 5.5cm 0 5.5cm},clip]{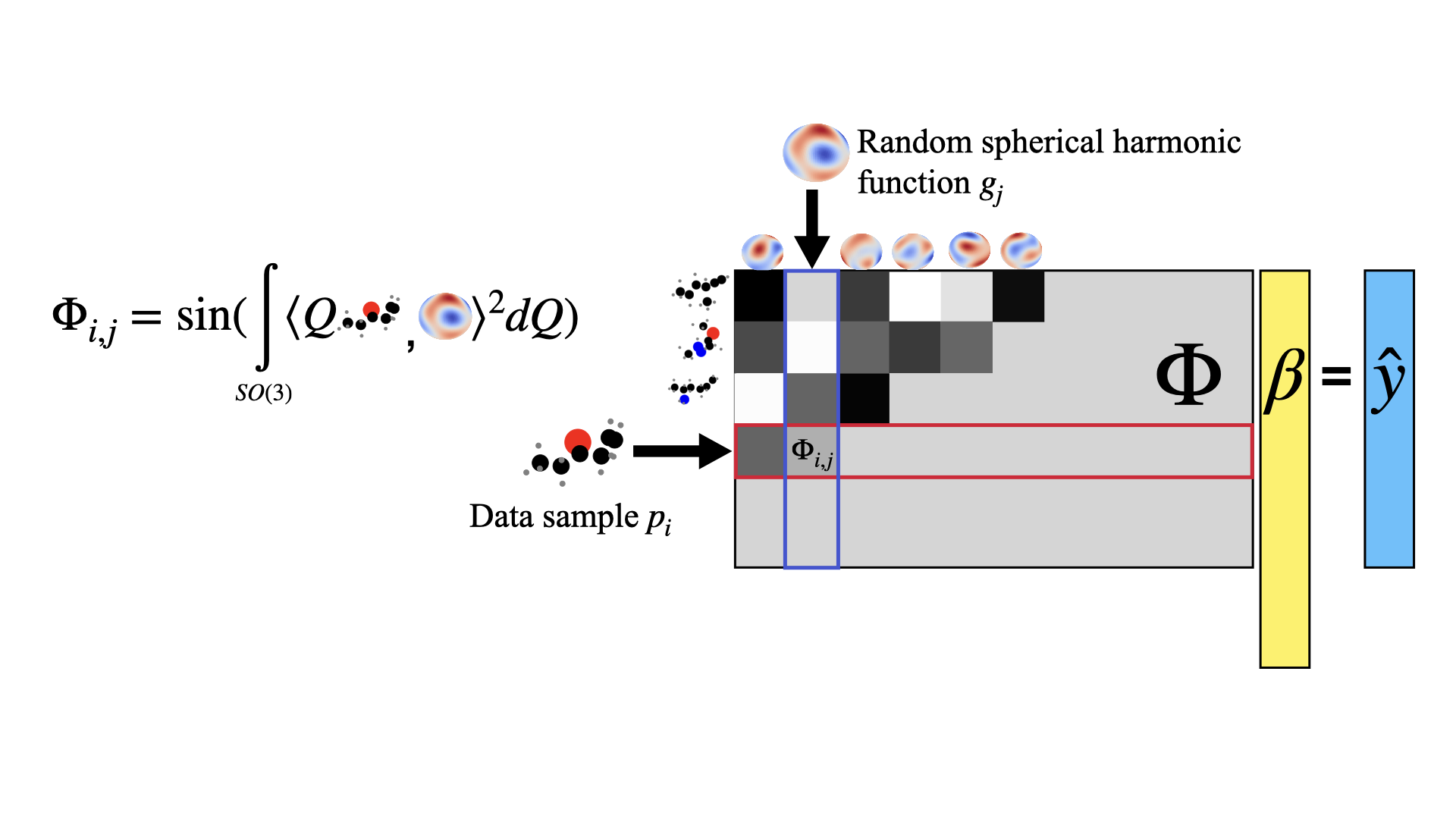}
    \caption[Method overview]{An overview of the rotation-invariant random features method. First we construct a feature matrix $\Phi$, defined in \cref{eq:random_feature_matrix_def}. The rows of $\Phi$ correspond to $p_i$, different samples in our training set, and the columns of $\Phi$ correspond to $g_j$, different random sums of spherical harmonic functions, which are defined in  \cref{eq:def_random_function_distribution}. The entries of the feature matrix are rotation-invariant random features. To compute the random features, we analytically evaluate the integral over all possible rotations of the data sample $p_i$. We describe our method for evaluating the random features in  \cref{sec:evaluating_random_features}. After constructing the feature matrix, we fit a set of linear weights $\beta$ and make predictions by evaluating the linear model $\Phi \beta = \hat{y}$.}
    \label{fig:method_overview}
\end{figure}

We propose a new method for learning rotation-invariant functions of point cloud data, which we call rotation-invariant random features. We extend the random features method of \citet{rahimi_random_2007} by deriving a version that is invariant to three-dimensional rotations and showing that it is fast to evaluate on point cloud data. 
The rotation invariance property of our features is clear from their definition, and computing the features only requires two simple results from the representation theory of $SO(3)$. Despite its simplicity, our method achieves performance near state-of-the-art on small-molecule energy regression and 3D shape classification. 
The strong performance on both tasks gives evidence that our method is general-purpose. 
An overview of our method can be found in \cref{fig:method_overview}. 

Our experiments in \cref{sec:QM7_experiments} show that our model is an order of magnitude faster than state-of-the-art kernel methods when predicting on new samples for energy regression tasks. Thus we conclude our model is a promising candidate for tasks which require real-time rotation-invariant predictions. 
Minimizing prediction latency is a growing concern for many machine learning methods; there are multiple applications requiring low-latency predictions on point clouds. 
Using machine learning predictions for real-time system control places hard requirements on prediction latencies. 
For example, trigger algorithms in the ATLAS experiment at the CERN Large Hadron Collider require prediction latencies ranging from $2 \mu s$ to $40 ms$ at different stages of the event filtering hierarchy \citep{ATLAS_ATLAS_2008}. The input to these trigger algorithms is a particle jet, which can be interpreted as a point cloud in four-dimensional spacetime. The majority of data-driven trigger algorithms either impose rotational invariance in phase space \citep{komiske_energy_2018,thaler_identifying_2011} or enforce invariance to the Lorentz group, which includes three-dimensional rotations and relativistic boosts \citep{bogatskiy_pelican_2022,gong_an-efficient_2022}. 
These definitions of invariance are compatible with our framework, with slight adjustments.
Another use-case for low-latency models is as replacements for force fields in molecular dynamics simulations \citep{gilmer_neural_nodate}. 
These calculations make serial subroutine calls to force field models, meaning latency improvements of these force field models are necessary to speed up the outer simulations. Other rotation-invariant machine learning models either rely on deep computational graphs, which provide a fundamental limit to speedup via parallelization, or they rely on kernel methods \citep{christensen_fchl_2020} which have prediction-time complexity linear in the number of training samples.

\subsection{Contributions}
\begin{itemize}
    \item We derive a rotation-invariant version of the standard random features approach presented in \citet{rahimi_random_2007}. By using simple ideas from the representation theory of $SO(3)$, our rotation-invariant random feature method is easy to describe and implement, and requires minimal expert knowledge in its design (\cref{sec:our_method}).
    \item We show with experiments that our rotation-invariant random feature method outperforms or matches the performance of general-purpose equivariant architectures optimized for small-molecule energy regression tasks. In particular, we outperform or match the performance of Spherical CNNs \citep{cohen_spherical_2018} on the QM7 dataset and Cormorant \citep{anderson_cormorant_2019} on the QM9 dataset. We compare our test errors with state-of-the-art methods on the QM9 dataset to quantify the benefit of designing task-specific rotation invariant architectures (\cref{sec:energy_regression_experiments}).
    \item We show that our method makes predictions an order of magnitude faster than kernel methods at test time (\cref{sec:QM7_experiments}).
    \item We show the rotation-invariant random feature method is general-purpose by achieving low test errors using the same method on a different task, shape classification on the ModelNet40 benchmark dataset (\cref{sec:ModelNet40_experiments}).
    \item We make our code publicly available at \\ \url{https://github.com/meliao/rotation-invariant-random-features}.
\end{itemize}

%% file: related_work.tex
\section{Related Work} 
\label{sec:related_work}
In this section, we describe two major types of rotation-invariant prediction methods. The first class of methods extracts rotation-invariant features using expert chemistry knowledge. The second class of methods uses deep neural networks to build general-purpose invariant architectures. Finally, we briefly survey the random features work that inspire our method. 

\paragraph{Descriptors of Atomic Environments} 
In the computational chemistry community, significant effort has been invested in using physical knowledge to generate ``descriptors of atomic environments,'' or rotation-invariant feature embeddings of molecular configurations. One such feature embedding is the Coulomb matrix, used by \citet{montavon_learning_nodate}.
In \citet{rupp_fast_2012}, the sorted eigenvalues of the Coulomb matrix are used as input features. 
The sorted eigenvalues are invariant to rotations of the molecule and give a more compact representation than the full Coulomb matrix. 
Noticing that the Coulomb matrix can relatively faithfully represent a molecule with only pairwise atomic interactions has been an important idea in this community. 
Many methods have been developed to generate descriptions of atomic environments by decomposing the representation into contributions from all atomic pairs, and other higher-order approximations consider triplets and quadruplets of atoms and beyond \citep{christensen_fchl_2020,bartok_representing_2013,kovacs_linear_2021,drautz_atomic_2019}.\footnote{Many general-purpose invariant methods follow this design as well. Cormorant \citep{anderson_cormorant_2019} has internal nodes which correspond to all pairs of atoms in the molecule, and Spherical CNNs \citep{cohen_spherical_2018} render spherical images from molecular point clouds using ideas from pairwise electrostatic (Coulombic) repulsion.} 
The FCHL19 \citep{christensen_fchl_2020} method creates a featurization that relies on chemical knowledge and is optimized for learning the energy of small organic molecules. The FCHL19 feature embedding depends on 2-body and 3-body terms. 
There are multiple different approaches, such as Smooth Overlap of Atomic Potentials (SOAP) \citep{bartok_representing_2013} and  Atomic Cluster Expansion (ACE) \citep{drautz_atomic_2019, kovacs_linear_2021} which first compute a rotation-invariant basis expansion of a chemical system, and then learn models in this basis expansion. 
In  \cref{sec:correspondance_with_ACE}, we show that our proposed method can be interpreted as using a highly-simplified set of ACE basis functions to learn a randomized nonlinear model. 

Deep learning is an effective tool in computational chemistry as well. Task-specific deep neural network architectures for learning molecular properties of small organic molecules outperform other approaches on common energy learning benchmarks.
Multiple methods \citep{schutt_schnet_2018, unke_PhysNet_2019} design neural networks to take atomic charges and interatomic distances as input. 
These methods use highly-optimized neural network architectures, including specially-designed atomic interaction blocks and self-attention layers. 
OrbNet \citep{qiao_orbnet_2020}, another neural network method, takes the output of a low-cost density functional theory calculation as rotation-invariant input features and uses a message-passing graph neural network architecture. 

\paragraph{General-Purpose Invariant Architectures}
While many general-purpose invariant architectures operate directly on point clouds, the first examples of such architectures, Spherical CNNs \citep{cohen_spherical_2018,esteves_learning_2018,kondor_clebsch-gordan_2018} operate on ``spherical images'', or functions defined on the unit sphere. These networks define spherical convolutions between a spherical image and a filter. These convolutions are performed in Fourier space, and the input spherical image is transformed using a fast Fourier transform on the unit sphere. Much like a traditional convolutional neural network, these networks are comprised of multiple convolutional layers in series with nonlinearities and pooling layers in between. To accommodate multiple different types of 3D data, including 3D point clouds, these methods take a preprocessing step to render the information of a 3D object into a spherical image.

Other methods operate on point clouds directly. SPHNet \citep{poulenard_effective_2019} computes the spherical harmonic power spectrum of the input point cloud, which is a set of rotation-invariant features, and passes this through a series of point convolutional layers. Cormorant \citep{anderson_cormorant_2019} and Tensor Field Networks \citep{thomas_tensor_2018} both design neural networks that at the first layer have activations corresponding to each point in the point cloud. All activations of these networks are so-called ``spherical tensors,'' which means they are equivariant to rotations of the input point cloud. To combine the spherical tensors at subsequent layers, Clebsch-Gordan products are used. Many other deep neural network architectures have been designed to operate on point clouds, but the majority have not been designed to respect rotational invariance. A survey of this literature can be found in \citet{guo_deep_2021}.

There are many ways to categorize the broad field of group-invariant deep neural network architectures, and we have chosen to form categories based on the input data type. Another helpful categorization is the distinction between \textit{regular} group-CNNs and \textit{steerable} group-CNNs introduced by \citet{cohen_intertwiners_2018}. In this distinction, \textit{regular} group-CNNs form intermediate representations that are scalar functions of the sphere or the rotation group, and examples of this category are \citet{cohen_spherical_2018,kondor_clebsch-gordan_2018,poulenard_effective_2019}. \textit{Steerable} group-CNNs form intermediate representations that are comprised of ``spherical tensors'', and examples of this category are \citet{anderson_cormorant_2019,weiler_3d_2018,thomas_tensor_2018}.

More recently, some works have investigated methods that endow non-invariant point cloud neural network architectures with a group-invariance property, either through a modification of the architecture or a data preprocessing step. \citet{xiao_endowing_2020} suggests using a preprocessing step that aligns all point clouds in a canonical rotation-invariant alignment; \citet{puny_frame_2022} expands this idea to general transformation groups. After the preprocessing step, any point cloud neural network architecture can be used to make rotation-invariant predictions of the input. Vector Neurons \citep{deng_vector_2021} endows neural network architectures with rotation equivariance by using three-dimensional vector activations for each neuron in the network and cleverly modifying the equivariant nonlinearities and pooling. 
These methods have been successfully applied to dense point clouds that arise in computer vision problems such as shape classification and segmentation. Molecular point clouds are qualitatively different; they are quite sparse, and the Euclidean nearest neighbor graph does not always match the graph defined by molecular bond structure. Because of these differences, the alignment methods and network architectures designed for solving shape classification and segmentation tasks are not \textit{a priori} good models for learning functions of molecular point clouds. To the best of our knowledge, they have not been applied successfully in molecular property prediction tasks.
Finally, \citet{villar_scalars_nodate} show that any rotation-invariant function of a point cloud can be expressed as a function of the inner products between individual points; this insight reduces the problem of designing rotation-invariant models to one of designing permutation-invariant architectures.

\paragraph{Random Features} 
Random Fourier features \citep{rahimi_random_2007} are an efficient, randomized method of approximating common kernels. For kernel methods, prediction time scales linearly with the size of the dataset. Random Fourier feature methods require time linear in $d$, an approximation parameter. The approximation works by drawing a random vector $\xi$ and defining a low-dimensional feature embedding $\varphi(x; \xi)$, called a random Fourier feature. The random Fourier features approximate a shift-invariant kernel $k(x, x') = k(x - x')$ in the sense that the inner product of two random Fourier feature evaluations is an unbiased estimate of the kernel:
\begin{equation*}
    \mathbb E_{\xi}\left[\langle \varphi(x; \xi), \varphi(x'; \xi) \rangle \right] = k(x, x')
\end{equation*}
This approximation holds when the random vectors $\xi$ are drawn from the Fourier transform of the kernel $k$ and the random Fourier features have the following functional form:
\begin{equation*}
    \varphi(x; \xi) = [\cos(\langle x, \xi \rangle), \ \ \sin(\langle x, \xi \rangle)]^\top
\end{equation*}
To reduce the variance of this approximation, one can draw $d$ such random vectors $\{ \xi_1, ..., \xi_d \}$ and concatenate the resulting random features. As a practical method, \cite{rahimi_random_2007} proposes building a feature matrix $\Phi \in \R^{n \times d}$ by drawing $d$ different random Fourier features $\varphi( \ \cdot \ ; \xi_j)$ and evaluating the random Fourier features at each of the $n$ samples $x_i$ in the training dataset:
\begin{equation*}
    \Phi_{i, j} = \varphi(x_i; \xi_j)
\end{equation*}
Once the feature matrix is formed, a  linear model is trained to fit a response vector $y$ using regularized linear regression, such as ridge regression:
\begin{equation}
    \argmin_\beta \| \Phi \beta - y \|_2^2 + \lambda \| \beta \|^2_2
    \label{eq:random_features_ridge_regression_setup}
\end{equation}
Experiments in \citet{rahimi_random_2007} show this is an effective method for fitting data, even when $d \ll n$. When $d \ll n$, the model can be evaluated much faster than the original kernel. 

Follow-up work, including \citet{rahimi_weighted_2008}, suggested that interpreting random feature methods as kernel approximators is not necessary. This work uses random nonlinear features with varying functional forms, including random decision stumps and randomly-initialized sigmoid neurons $\sigma(\langle \xi, x \rangle)$. These random nonlinear features form effective models for diverse types of data. In  \cref{sec:our_method}, we introduce random features similar to those in \citet{rahimi_random_2007}. Our method does not approximate any explicit kernel, and it is designed to be invariant to any rotation of the input data, so we call our method rotation-invariant random features. 

Finally, the notion of group-invariant random feature models also appears in \citet{mei_learning_2021} as a technical tool to understand the sample complexity benefit of enforcing group invariances in overparameterized models. In this work, the invariances considered are transformations on one-dimensional signals that arise from cyclic and translation groups.

%% file: rotational_invariance.tex
\section{Rotational Invariance and Spherical Harmonics}
In this section, we will introduce rotational invariance and the spherical harmonics. In our method, we use the definitions presented in this section and only two simple facts about the spherical harmonics and rotations, which we list at the end of this section.
We say a function $f$ mapping a point cloud $p$ to an element of the vector space $\mathcal{Y}$ is rotation-equivariant if there exists some group action on the vector space $\mathcal{Y}$ such that 
\begin{equation*}
  f(Q \circ p) = Q \circ f(p) \ \ \forall Q \in SO(3)
\end{equation*}
Intuitively, this means that when an input to $f$ is rotated, $f$ preserves the group structure and the output changes predictably. We say that a function $f$ is rotation-invariant if the group action on the output vector space $\mathcal{Y}$ is the identity:
\begin{equation}
  f(p) = f(Q \circ p ) = f(\{ Qx_1, Qx_2, \hdots, Qx_N \}) \ \ \ \forall Q \in SO(3)
  \label{eq:invariance_condition}
\end{equation}

Finally, our goal is to learn a function over point clouds that does not change when any rotation is applied to the point cloud. Formally, given some distribution $\mathcal{D}$ generating pairs of data $(p, y)$ our learning goal is to find 
\begin{equation*}
  f^* = \argmin_{f} \mathbb E_{(p, y) \sim \mathcal{D}}\left[ l(f(p), y) \right] 
\end{equation*} 
where $l$ is a classification or regression loss depending on the task and the minimization is over all functions that satisfy a rotational invariance constraint \cref{eq:invariance_condition}.

\subsection{Spherical Harmonics}
\label{sec:spherical_harmonics}
When decomposing a periodic function $f: [0, 2 \pi] \to \R$, a natural choice of basis functions for the decomposition is the complex exponentials $e^{imx}$. In this basis, $f(x)$ may be expressed as $f(x) = \sum_m a_m e^{i m x}$ where $a_m = \langle f(x), e^{imx} \rangle$.
When decomposing a function on the unit sphere $S^2$, a similar set of basis functions emerges, and they are called the spherical harmonics. The complex exponentials are indexed by a single parameter $m$, and because the spherical harmonics span a more complicated space of functions, they require two indices, $\ell$ and $m$. By convention, the indices are restricted to $\ell \geq 0$ and $- \ell \leq m \leq \ell$. 
A single spherical harmonic function $Y^{(\ell)}_m$ maps from the unit sphere $S^2$ to the complex plane $\C$. For a function $f(x) : S^2 \to \R$ we can compute an expansion in the spherical harmonic basis:
\begin{align*}
  f(x) &= \sum_{m, \ell} a_m^{(\ell)} Y_m^{(\ell)}(x) \\
  a_m^{(\ell)} &= \langle f(x), Y_m^{(\ell)}(x) \rangle
\end{align*}

In designing our method, we use two simple facts about spherical harmonics: 
\begin{itemize}
  \item If one has evaluated the spherical harmonics of some original point $x \in S^2$ and wants to evaluate those spherical harmonics at a new rotated point $Qx$ for some rotation matrix $Q$, the rotated evaluation is a linear combination of the un-rotated spherical harmonics at the same index $\ell$:
  \begin{equation}
    Y_m^{(\ell)}(Qx) = \sum_{m' = -\ell}^\ell Y_{m'}^{(\ell)}(x) D^{(\ell)}(Q)_{m,m'}
  \end{equation}
  Here $D^{\ell}(Q)$ is a Wigner-D matrix; it is a square matrix of size $(2\ell + 1 \times 2\ell + 1)$.
  \item Evaluating the inner product between two elements of the Wigner D-matrices is simple. In particular, when $dQ$ is the uniform measure over $SO(3)$, we have the following expression:
  \begin{equation}
    \int_{SO(3)} D^{(\ell_1)}(Q)_{m_1,k_1} D^{(\ell_2)}(Q)_{m_2,k_2} dQ = (-1)^{m_1-k_1} \frac{8 \pi^2}{2 \ell + 1} \delta_{-m_1, m_2} \delta_{-k_1, k_2} \delta_{\ell_1, \ell_2}
  \end{equation}
  This is a corollary of Schur's lemma from group representation theory, and we discuss this fact in \cref{sec:Wigner_D_matrix_proof}.
\end{itemize}

%% file: our_method.tex
\section{Rotation-Invariant Random Features}\label{sec:our_method}

In our setting, an individual data sample is a point cloud $p_i = \{ x_{i,1}, \hdots, x_{i,{N_i}} \}$ with points $x_{i,j} \in \R^3$. 
We model the data as a function: $p_i(x) = \sum_{j=1}^{N_i} \delta(x - x_{i,j})$ where $\delta(\cdot )$ is a delta function centered at 0. 
We can then use a functional version of the random feature method where our data $p_i$ is a function, $g$ is a random function drawn from some pre-specified distribution, and $\langle \cdot, \cdot \rangle$ is the $L^2$ inner product:
\begin{equation}
  \tilde\varphi(p_i; g) = \sin \left( \langle p_i, g \rangle \right)
  \label{eq:first_try_random_feature}
\end{equation}
We want the random feature to remain unchanged after rotating the point cloud,
but \cref{eq:first_try_random_feature} does not satisfy this. For general functions $g$, $\langle Q \circ p_i , g \rangle \neq \langle p_i, g \rangle $.
We achieve rotational invariance by defining the following rotation-invariant random feature: 
\begin{equation}
  \varphi(p_i; g) = \sin \left( \int_{SO(3)} \langle Q \circ p_i, g \rangle^2 dQ \right)
  \label{eq:invariant_random_features_integral}
\end{equation}
We ``symmetrize'' the inner product by integrating over all possible orientations of the point cloud $p_i$, eliminating any dependence of $\varphi( \cdot; g_j)$ on the data's initial orientation. Because $\varphi( \cdot; g_j)$ does not depend on the initial orientation of the data, it will not change if this initial orientation changes, i.e., if $p_i$ is rotated by $Q$. To build a regression model, we construct a feature matrix 
\begin{equation}
  \Phi_{i, j} = \varphi(p_i; g_j)
  \label{eq:random_feature_matrix_def}
\end{equation}
and fit a linear model to this feature matrix using ridge regression, as in \cref{eq:random_features_ridge_regression_setup}.

\subsection{Evaluating the Random Features}
\label{sec:evaluating_random_features}
In this section, we describe how to efficiently evaluate the integral in \cref{eq:invariant_random_features_integral}. $SO(3)$ is a three-dimensional manifold and the integral has a nonlinear dependence on the data, so methods of evaluating this integral are not immediately clear. However, representation theory of $SO(3)$ renders this integral analytically tractable and efficiently computable. The main ideas used to evaluate this integral are exploiting the linearity of inner products and integration, and choosing a particular distribution of random functions $g$. 
First, we observe that because the rotated data function $Q \circ p_{i}$ is the sum of a few delta functions, we can expand the inner product as a sum of evaluations of the random function $g$:
\begin{equation}
  \langle Q \circ p_{i}, g \rangle = \sum_{j=1}^{N_i} g(Qx_{i,j})
\end{equation}
Expanding the inner product and the quadratic that appear in \cref{eq:invariant_random_features_integral} and using the linearity of the integral, we are left with a sum of simpler integrals:
\begin{equation}
  \int_{SO(3)} \langle Q \circ p_i, g \rangle^2 dQ = \sum_{j_1, j_2 = 1}^{N_i} \int_{SO(3)} g(Qx_{i, j_1})g(Qx_{i, j_2}) dQ
  \label{eq:integral_expansion_1}
\end{equation}
Next, we choose a particular distribution for $g$ which allows for easy integration over $SO(3)$. We choose to decompose $g$ as a sum of randomly-weighted spherical harmonics with maximum order $L$ and $K$ fixed radial functions: 
\begin{equation}
  g(x) = \sum_{k=1}^K \sum_{\ell = 0}^L \sum_{m=- \ell}^\ell w^{(\ell)}_{m,k} Y^{(\ell)}_m(\hat{x}) R_k(\| x \|)
  \label{eq:def_random_function_distribution}
\end{equation}
where $w^{(\ell)}_{m,k}$ are random weights drawn from a Gaussian distribution $\mathcal{N}(0, \sigma^2)$, $\hat x = \frac{x}{\|x \|}$, and $R_k: \R_+ \to \R$ is a radial function. Two examples of such radial functions are shown in  \cref{fig:radial_funcs}. This definition of random functions gives us a few hyperparameters; $L$, the maximum frequency of the spherical harmonics; $\sigma$, the standard deviation of the randomly-drawn weights; and the choice of radial functions. In \cref{sec:supplementary_figures}, we show that the performance of our method is most sensitive to $\sigma$.

By repeated application of the linearity of the integral and the two facts about the spherical harmonics mentioned in  \cref{sec:spherical_harmonics}, we are able to write down a closed-form expression for the integral on the left-hand side of \cref{eq:integral_expansion_1}. 
We include these algebraic details in \cref{sec:full_integration_details}. Computing this integral has relatively low complexity; it requires evaluating a table of spherical harmonics for each point $x_{i, j}$ up to maximum order $L$, and then performing a particular tensor contraction between the array of spherical harmonic evaluations and the array of random weights. This tensor contraction has complexity $O(N^2L^3K^2)$. 

%% file: experiments.tex
\section{Experiments}
\label{sec:experiments}
We conduct multiple experiments comparing our method with other landmark rotation-invariant machine learning methods. We find that for predicting the atomization energy of small molecules, our method outperforms neural networks in the small dataset setting, and we have competitive test errors in the large dataset setting.
Our method is an order of magnitude faster than competing kernel methods at test time. 
We also find that our method performs competitively on a completely different task, 3D shape classification. 
We perform these experiments to show that our method can be used as a fast, simple, and flexible baseline for rotation-invariant prediction problems on 3D point cloud data. 

\subsection{Small-Molecule Energy Regression}
\label{sec:energy_regression_experiments}
A common target for machine learning in chemistry is the prediction of a potential energy surface, which maps from 3D atomic configurations to the atomization energy of a molecule.
Large standardized datasets such as QM7 \citep{blum_970_2009, rupp_fast_2012} and QM9 \citep{ruddigkeit_enumeration_2012,ramakrishnan_quantum_2014} offer a convenient way to test the performance of these machine learning models across a wide chemical space of small organic molecules. 
Both datasets contain the 3D atomic coordinates at equilibrium and corresponding internal energies. 
The 3D coordinates and molecular properties are obtained by costly quantum mechanical calculations, so there may be settings which require high-throughput screening where an approximate but fast machine learning model may provide an advantageous alternative to classical methods.
To adapt our method to this task, we make two small changes. 

\paragraph*{Element-Type Encoding} The rotation-invariant random feature method defined above works for general, unlabeled point clouds.
However, in the chemistry datasets mentioned above, we are given more information than just the 3D coordinates of particles in the molecule: the particles are individual atoms, and we know their element type. 
Incorporating the element type of individual atoms in a machine learning method is crucial for accurate prediction. Different elements interact in quantitatively different ways, as predicted by the laws of gravitation and electrostatic repulsion, and they interact in qualitatively different ways due to their electron configurations. 

We use an element-type encoding method inspired by FCHL19 \citep{christensen_fchl_2020} and ACE \citep{kovacs_linear_2021}. 
To construct this element-type encoding, we look at the local view of a molecule created by centering the atomic coordinates at a given atom, separate the atoms into different point clouds for each element type, and compute one random feature per element type. 
We then repeat this procedure for all elements of a given type and sum their feature vectors. 

Expressed mathematically, we are given a sample $p_i = \{ x_{i,1} , \hdots, x_{i,N_i} \}$ with charges $\{ c_{i,1} , \hdots, c_{i,N_i} \}$. 
Let $p_i^{(c_j)}$ be the collection of atoms with charge $c_j$, and let $p_i^{(c_j)} - x_{i,h}$ denote the point cloud of atoms in $p_i$ with charge $c_j$ centered at point $x_{i,h}$. In this notation,  $p_i^{(c_j)} - x_{i, h}$ specifies a point cloud, which we can treat as unlabeled because they all have the same charge. As before, a random feature is denoted $\varphi( \cdot; g_j)$. Then for all possible element-type pairs $(c_1, c_2)$, we compute individual entries in our feature matrix
\begin{align*}
    \Phi_{i,j'} = \sum_{h: \  c_{i_h} = c_1} \varphi \left( p_i^{(c_2)} - x_{i,h} ; g_j \right)
\end{align*}
The column index $j'$ depends on $j, c_1$, and $c_2$.

\begin{figure}
  \begin{subfigure}[b]{0.5\textwidth}
    \centering
    \includegraphics[width=\textwidth]{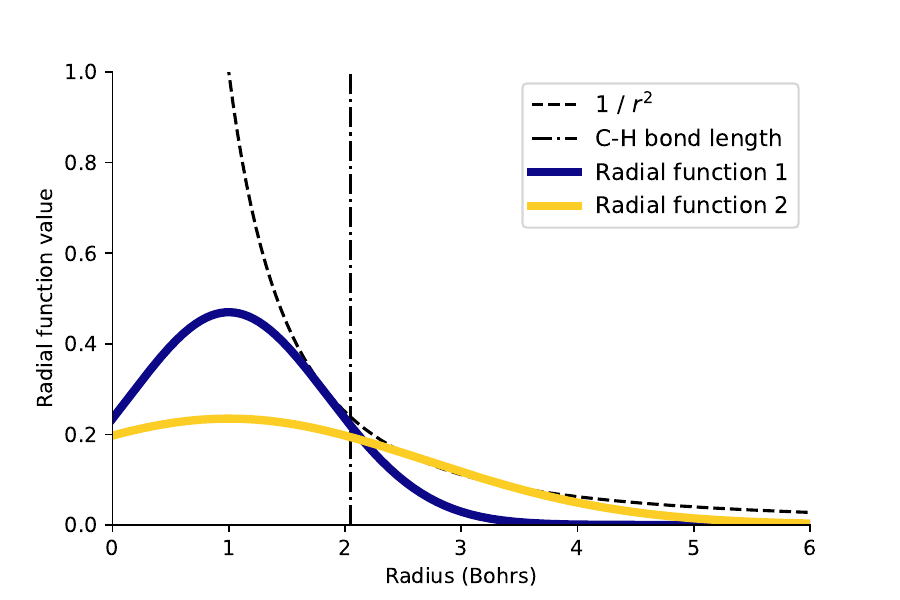}
    \caption{Radial functions for small-molecule energy regression}
    \label{fig:radial_funcs_a}
  \end{subfigure}
  \begin{subfigure}[b]{0.5\textwidth}
    \centering
    \includegraphics[width=\textwidth]{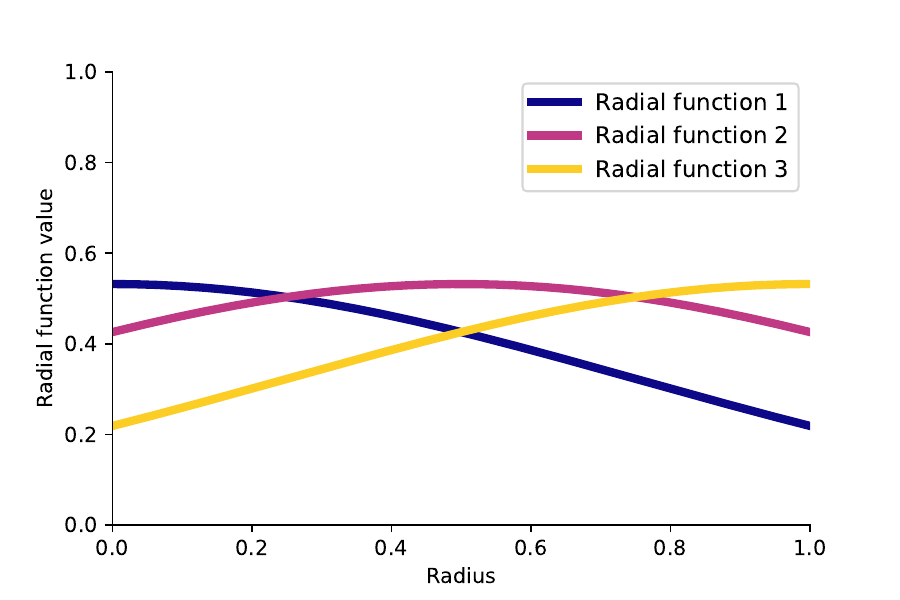}
    \caption{Radial functions for 3D shape classification}
    \label{fig:radial_funcs_b}
  \end{subfigure}
  \caption[Radial functions used in our method]{Our rotation-invariant random feature method requires simple user-defined radial functions. \cref{fig:radial_funcs_a} shows the radial functions used in our small-molecule energy regression experiments. We include $\frac{1}{r^2}$ and the average Carbon-Hydrogen bond length to give context to the horizontal axis. \cref{fig:radial_funcs_b} shows the radial functions used in our 3D shape classification experiments. The 3D shapes in the benchmark dataset are normalized to fit inside the unit sphere.}
  \label{fig:radial_funcs}
\end{figure}

\paragraph*{Radial Functions} 
We choose to parameterize our set of random functions as randomly-weighted spherical harmonics with fixed radial functions. 
The choice of radial functions appears as a design decision in many rotation-invariant machine learning methods. For example, radial functions appear as explicit design choices in \cite{kovacs_linear_2021,christensen_fchl_2020,poulenard_effective_2019} and implicitly in the design of \cite{cohen_spherical_2018}.
This design choice is often paramount to the empirical success of these methods. 
FCHL19 \citep{christensen_fchl_2020} optimize their radial functions over multiple hyperparameters and carefully balance multiplicative terms including log-normal radial functions, polynomial decay,  and soft cut-offs. 
The ACE models in \cite{kovacs_linear_2021} use a set of orthogonal polynomials defined over a carefully-chosen subset of the real line, a physically-motivated spatial transformation, and use extra ad-hoc radial functions that control the behavior of extremely nearby points.
For our radial functions, we use two Gaussians, both centered at 1, with width parameters chosen so the full widths at half maximum are 2 and 4 respectively. Our radial functions are shown in  \cref{fig:radial_funcs}.

\subsubsection{QM7 Atomization Energy Regression}
\label{sec:QM7_experiments}
The QM7 dataset contains 7,165 small molecules with element types H, C, N, O, S and a maximum of 7 heavy elements.
We compare with two lightweight methods, FCHL19 and Spherical CNNs, that show learning results on the QM7 dataset. We reproduce their original training methods and compare test errors in \cref{tab:QM7_results}. Notably, our rotation-invariant random feature method has average errors half of those of Spherical CNNs while being faster to train. Our best-performing model uses 2,000 random features. With the element-type encoding described above, this corresponds to 50,000 trainable parameters. 

In this experiment, we use a training dataset of 5,732 samples and a test set of size 1,433. For our method and FCHL19, we use 90\% of the training set to train the models and the remaining 10\% as a validation set for hyperparameter optimization. To train the FCHL19 models, we use the validation set to optimize over Gaussian kernel widths and $L^2$ regularization parameters. In our method, we search over the number of random features and $L^2$ regularization parameters. We solve our ridge regression problem by taking a singular value decomposition of the random feature matrix, and then constructing a solution for each regularization level.

In addition to our ridge regression models, we also experiment using our random features as an input to deep neural networks. We are unable to find deep neural network models with reasonable prediction latencies that outperformed our ridge regression model (\cref{sec:extra_latency_experiments}). We also experimented with a model which combined a rotation-invariant PCA alignment \citep{xiao_endowing_2020,puny_frame_2022} with three-dimensional random features, but the alignments produced by this method were uninformative, and the model failed to meet basic baselines.

\newcommand{\cx}{0.25\textwidth}
\newcommand{\greencheck}{\textcolor{green}{\cmark}}
\newcommand{\redx}{\textcolor{red}{\xmark}}
\begin{table}
    \centering
    \begin{tabular}{p{\cx}llll}\toprule
        \textbf{Method} & \textbf{General-}  & \textbf{Mean Absolute} & \textbf{Train Time (s)} & \textbf{Train Device} \\
        & \textbf{Purpose}  & \textbf{Error (eV)} &  &  \\ \midrule 
        \multirow{2}{\cx}{Spherical CNNs \\ \small{\citep{cohen_spherical_2018}}}  & \greencheck & 0.1565   & 546.7        & single GPU \\ \\
        \multirow{2}{\cx}{Random Features \\ \small{(Ours)}} & \greencheck & 0.0660 $\pm$ 0.00275 & 203.4 & 24 CPU cores \\ \\ 
        \multirow{2}{\cx}{FCHL19 \\ \small{\citep{christensen_fchl_2020}}} & \redx       &  0.0541 & 260.7 & 24 CPU cores \\ \\
        \bottomrule 
    \end{tabular}
    \caption[Comparison of atomization energy regression results on the QM7 dataset]{Test error and training time on the QM7 dataset. We report the mean absolute error on the test set for all methods and the standard error of the mean for our method. We discuss this experiment in \cref{sec:QM7_experiments}.}
    \label{tab:QM7_results}
\end{table}

The space of small organic drug-like molecules is extremely large. The GDB-13 dataset \citep{blum_970_2009} contains nearly one billion reference molecules, even after filtering for molecule size and synthesis prospects.
Most of the machine learning methods approximating potential energy surfaces are considered fast approximate replacements for costly quantum mechanical calculations. To screen the extremely large space of small organic drug-like molecules, these methods require high prediction throughput, which can be achieved by exploiting parallelism in the models' computational graphs and readily-available multicore CPU architectures. When parallelism inside the model is not available, the set of candidate molecules can be divided and data parallelism can be used to increase screening throughput. 
However, we consider another setting where experimental samples need to be analyzed in real time to make decisions about ongoing dynamic experiments. In this setting, prediction latency is paramount. 

In the low-latency setting, our method can gracefully trade off prediction latency and prediction error by tuning the number of random features used in a given model. Importantly, all training samples are used to train the model. FCHL19 is a kernel method, so the time required to predict a new data point scales linearly with the number of training samples used. Thus, FCHL19 can only improve test latency at the cost of using fewer training samples and incurring higher test errors. 

\begin{figure}
  \begin{minipage}[b]{0.5\textwidth}
    \includegraphics[width=\textwidth,trim={0 0 15cm 0},clip]{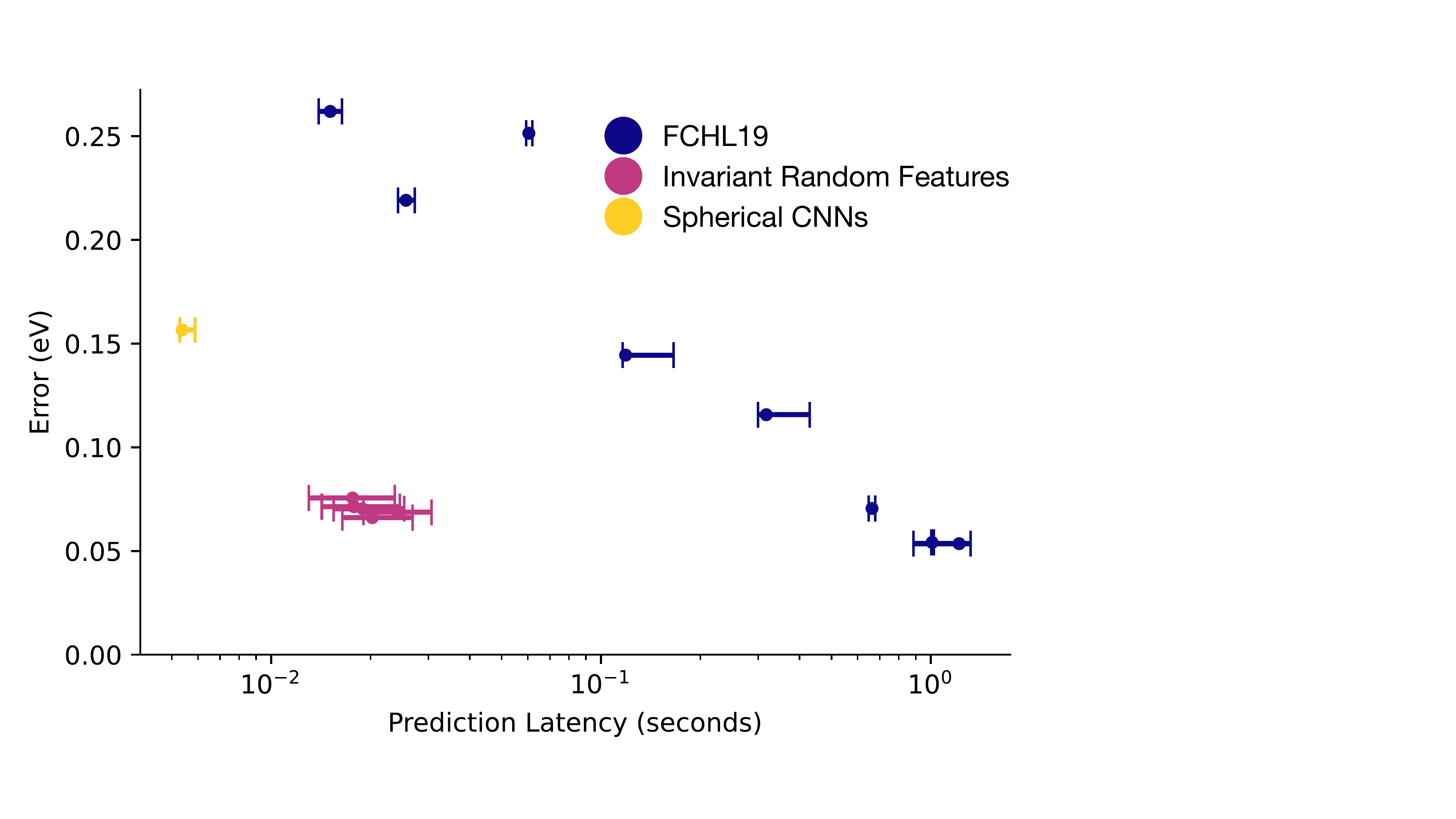}
    \vspace{-3cm}
  \end{minipage} 
  \begin{minipage}[b]{0.5\textwidth}
    \vspace{-3cm}
    \resizebox{\textwidth}{!}{\begin{tabular}{lp{1cm}p{2cm}p{1.5cm}l}
      \toprule
      \textbf{Method} & \textbf{Model Size} & \textbf{Prediction Latency (s)} & \textbf{Train time (s)} & \textbf{Error (eV)} \\
      \midrule
      FCHL19 & 100 & 0.0151 & 0.4 & 0.2620 \\
      FCHL19 & 200 & 0.0256 & 0.8 & 0.2191 \\
      FCHL19 & 400 & 0.0604 & 2.3 & 0.2514 \\
      FCHL19 & 800 & 0.1188 & 9.1 & 0.1444 \\
      FCHL19 & 1600 & 0.3170 & 54.0 & 0.1157 \\
      FCHL19 & 3200 & 0.6635 & 258.5 & 0.0705 \\
      FCHL19 & 5000 & 1.0096 & 260.7 & 0.0541 \\
      FCHL19 & 5732 & 1.2191 & 326.1 & 0.0535 \\
      Random Features (Ours) & 250 & 0.0177 & 61.6 & 0.0755 \\
      Random Features (Ours) & 500 & 0.0178 & 107.1 & 0.0714 \\
      Random Features (Ours) & 750 & 0.0189 & 157.8 & 0.0704 \\
      Random Features (Ours) & 1000 & 0.0203 & 203.4 & 0.0660 \\
      Random Features (Ours) & 2000 & 0.0242 & 431.4 & 0.0687 \\
      Spherical CNNs & N/A & 0.0054 & 546.7 & 0.1565\\
      \bottomrule
    \end{tabular}}
  \end{minipage}
  \caption[Comparison of test latency and test error on the QM7 dataset]{Invariant Random Features are faster at prediction time than FCHL19 and more accurate than Spherical CNNs. In this experiment, we compare the prediction latency and prediction error for models of different sizes evaluated on the QM7 dataset. For each method, we measure the latency of predicting each molecule on a held-out test set 5 times. In the figure, we plot the median prediction latencies with error bars spanning the $25^{th}$ and $75^{th}$ percentile measurements. The vertical axis of the figure is mean absolute error measured in eV on a held-out test set. The FCHL19 method is evaluated on 24 CPU cores using kernels with sample sizes 100, 200, 400, 800, 1600, 3200, 5000, and 5732. The rotation-invariant random features method is evaluated on 24 CPU cores with model sizes of 250, 500, 750, 1000, and 2000 random features. The Spherical CNNs method is evaluated on a single GPU. We discuss this experiment in \cref{sec:QM7_experiments}.}
  \label{fig:test_latency_vs_test_error}
\end{figure}

To explore the tradeoff between prediction latency and prediction error, we train random feature models and FCHL19 models of different sizes. We measure their prediction latencies and plot the results in \cref{fig:test_latency_vs_test_error}. We note that at similar error levels, FCHL19 models show prediction latencies that are an order of magnitude slower than ours. Spherical CNNs exhibit low test latency because their method is implemented for a GPU architecture, but their test errors are high, and the neural network approach does not admit obvious ways to achieve a tradeoff between latency and test error. The prediction latencies of FCHL19 and our random feature models depend on the number of atoms in the molecule, and we show this dependence in \cref{fig:latency_by_n_atoms}.

\subsubsection{QM9 Atomization Energy Regression}
\label{sec:QM9_experiments}
\newcommand\xxx{\par\hangindent1em\makebox[1em][l]{$\bullet$}}
\newcommand{\cw}{0.2\textwidth}
\begin{table}[h!]
  \centering
  \begin{tabular}{p{\cw}p{0.1\textwidth} p{0.2\textwidth} p{0.165\textwidth} p{0.2\textwidth} }\toprule
      \textbf{Method} & \textbf{General-Purpose} & \textbf{Model Type} & \textbf{Mean Absolute Error (eV)} & \textbf{Input Features}  \\ \midrule 
      \multirow{2}{\cw}{Cormorant \\ \small{\citep{anderson_cormorant_2019}}} & \greencheck & General-purpose rotation-invariant architecture 
      & 0.022 %
      & \xxx Charges \xxx Atomic locations  
      \\ \midrule
      \multirow{2}{\cw}{Random Features \\ \small{(Ours)}} & \greencheck & Ridge Regression 
      & 0.022 $\pm$ 4.45e-04 %
      & \xxx Charges \xxx Rotation-invariant random features 
      \\  \midrule
      \multirow{2}{\cw}{FCHL19 \\ \small{\citep{christensen_fchl_2020}}} & \redx & Kernel Ridge Regression 
      &  0.011 %
      & \xxx Charges \xxx Interatomic distances \xxx 3-body angles  \\ \midrule
      \multirow{2}{\cw}{SchNet \\ \small{\citep{schutt_schnet_2018}}} & \redx & Neural Network with atomic interaction blocks 
      & 0.014 %
      & \xxx Charges \xxx Interatomic  distances \\ \midrule 
      \multirow{2}{\cw}{PhysNet \\ \small{\citep{unke_PhysNet_2019}}} & \redx & Neural Network with attention layers 
      & 0.008 %
      & \xxx Charges \xxx Interatomic distances \\ \midrule
      
      \multirow{2}{\cw}{OrbNet \\ \small{\citep{qiao_orbnet_2020}}} & \redx & Message-Passing GNN 
      & 0.005 %
      & \xxx Mean-field DFT calculations  \\
      \bottomrule
  \end{tabular}
  \caption[Comparison of atomization energy regression results on the QM9 dataset]{Comparing large-scale models trained on QM9. FCHL19 \citep{christensen_fchl_2020} was trained on 75,000 samples from QM9 and the other methods were trained on 100,000 samples. 
  We report the mean absolute error on the test set for all methods and the standard error of the mean for our method. We discuss this experiment in \cref{sec:QM9_experiments}.
  }
  \label{tab:QM9_results}
\end{table}

The QM9 dataset contains 133,885 molecules with up to 9 heavy atoms C, O, N, and F, as well as Hydrogen atoms. This is a larger and more complex dataset than QM7, and because there is more training data, neural network methods perform well on this benchmark. For this dataset, we follow \cite{anderson_cormorant_2019, gilmer_neural_nodate} by constructing atomization energy as the difference between the internal energy at $0K$ and the thermochemical energy of a molecule's constituent atoms. We train a large-scale random feature model by taking 100,000 training samples from QM9 and generating 10,000 random features. We select an $L^2$ regularization parameter \textit{a priori} by observing optimal regularization parameters from smaller scale experiments on a proper subset of the QM9 training data.

We compare the performance of our model with the reported errors of FCHL19 \citep{christensen_fchl_2020} and other neural network methods. All methods considered in this comparison enforce rotational invariance in their predictions. 
The results of this comparison are shown in  \cref{tab:QM9_results}. 
Our rotation-invariant random feature method matches the performance of Cormorant, and it provides a very strong baseline against which we can quantify the effect of expert chemistry knowledge or neural network design.
OrbNet \citep{qiao_orbnet_2020} uses a graph neural network architecture, and their inputs are carefully chosen from the results of density functional theory calculations in a rotation-invariant basis. PhysNet \citep{unke_PhysNet_2019} uses an architecture heavily inspired by that of SCHNet \citep{schutt_schnet_2018}, and this iteration in model design resulted in almost a 50\% reduction in test error. 

Training models on large subsets of the QM9 dataset is difficult. FCHL19 requires 27 hours to construct a complete kernel matrix of size (133,855 $\times$ 133,855) on a compute node with 24 processors. The 27 hours does not include the time required to find the model's linear weights. 
The neural network methods require long training sequences on GPUs. Both Cormorant and PhysNet report training for 48 hours on a GPU.

When using a large sample size, our method is similarly difficult to train. 
We are able to construct a matrix of random features of size (100,000 $\times$ 250,000) in 27 minutes on a machine with 24 processors, but the matrix is highly ill-conditioned, and using an iterative method to approximately solve the ridge regression problem requires 76.5 hours.
Using an iterative method introduces approximation error which we did not encounter when performing experiments on the other datasets. 
We attribute some of the performance gap between our method and FCHL19 to this approximation error. 
Fortunately, solving large, dense, overdetermined ridge regression problems is an active area of research, and it is quite likely that applying methods from this area of research to our problem would result in a speed-up. We outline some promising approaches in \cref{sec:ridge_regression_appendix}.
We believe the conditioning of our problem is caused by the choice of element-type encoding, and we leave finding a new element-type encoding that produces well-conditioned feature matrices for future work.

\subsection{Shape Classification}
\label{sec:ModelNet40_experiments}
To show that our method is general-purpose, we test our same method on a completely different task, 3D shape classification. In particular, we consider multiclass classification on the ModelNet40 benchmark dataset.  
The ModelNet40 benchmark dataset \cite{zhirong_wu_3d_2015} is a set of computer-generated 3D models of common shapes, such as mugs, tables, and airplanes. 
We evaluate the performance of our model in three train/test settings, where the train and test sets either contain rotations about the $z$ axis, or arbitrary rotations in $SO(3)$.
There are 9,843 training examples and 2,468 test examples spread across 40 different classes. The shape objects are specified by 3D triangular meshes, which define a (possibly disconnected) object surface. 
To generate a point cloud from individual objects in this dataset, one must choose a sampling strategy and sample points from the surface of the object. 
We use the dataset generated by \cite{qi_pointnet_2017}, which samples 1,024 points on the mesh faces uniformly at random. The point clouds are then centered at the origin and scaled to fit inside the unit sphere. 

We solve the multiclass classification problem with multinomial logistic regression. More specifically, we optimize the binary cross entropy loss with L2 regularization to learn a set of linear weights for each of the 40 classes in the dataset. At test time, we evaluate the 40 different linear models and predict by choosing the class with the highest prediction score. 
We also slightly change the definition of our data function; for a point cloud $p_i = \{ x_{i,1}, ..., x_{i,{N_i}} \}$, we use a normalized data function $p_i(x) = \frac{1}{N_i} \sum_{j=1}^{N_i} \delta(x - x_{i,j})$ to eliminate any dependence on the number of points sampled from the surface of the shape objects. For this setting, we use radial functions that are three Gaussian bumps with centers at $0, 0.5$, and $ 1$, with width $\sigma=0.75$. 

We compare our method with another rotation-invariant method, SPHNet \citep{poulenard_effective_2019}. 
SPHNet is a multilayer point-convolutional neural network that enforces rotational invariance by computing the spherical harmonic power spectrum of the input point cloud. SPHNet's rotation-invariant method also requires a choice of radial functions, and they use two Gaussians with different centers and vary the width of the Gaussians at different layers of their network. The results of our comparison are in \cref{tab:Modelnet40_results}. Our method does not achieve near state-of-the-art results, but we conclude it provides a strong baseline on this challenging multiclass problem. We discuss the prediction latency of our method on the ModelNet40 dataset in \cref{sec:extra_latency_experiments}.

\begin{table}
  \centering
  \begin{tabular}{p{0.35\textwidth}p{0.2\textwidth}rrr}\toprule
      \textbf{Method} & \textbf{Rotation Invariant} & $\mathbf{z/z}$ & $\mathbf{SO(3)/SO(3)}$ & $\mathbf{z/SO(3)}$ \\ \midrule 
      Spherical CNNs \small{\citep{esteves_learning_2018}} & \greencheck & 0.889 & 0.869 & 0.786 \\
      SPHNet \small{\citep{poulenard_effective_2019}} & \greencheck & 0.789 & 0.786 & 0.779 \\ %
      Vector Neurons \small{\citep{deng_vector_2021}} & \greencheck & 0.902 & 0.895 & 0.895 \\
      Random Features \small{(Ours)} & \greencheck & 0.693 & 0.692 & 0.666 \\ %
      PointNet++ \small{\citep{qi_pointnetplusplus_2017}} & \redx & 0.918 & 0.850 & 0.284 \\
      \bottomrule
  \end{tabular}
  \caption[Comparison of shape classification results on the ModelNet40 dataset]{Test accuracy on the ModelNet40 shape classification benchmark task. Our method sets a strong baseline to compare against neural network methods, especially in the challenging $z/SO(3)$ train/test regime. We discuss this experiment in \cref{sec:ModelNet40_experiments}.
  }
  \label{tab:Modelnet40_results}
\end{table}

%% file: conclusion.tex
\section{Discussion and Future Work}
Our method is a simple, flexible, and competitive baseline for rotationally-invariant prediction problems on 3D point clouds. Rotation-invariant random features are simple to explain and implement, and using them in varied prediction tasks requires a minimal amount of design choices. Our method does not achieve state-of-the-art accuracy on any prediction task, but it provides a competitive baseline that allows us to begin to quantify the effect of neural network models and expert chemistry knowledge in methods that enforce rotational invariance. 
In particular, we find that for molecular property prediction tasks, the inductive bias of rotation invariance is enough to provide very strong performance with our general-purpose random feature model. For 3D shape classification, we find that the inductive bias of rotation invariance provides a non-trivial baseline, but the gap between our model and neural networks is larger.

Our method shows promise in settings where low prediction latency is desired. 
In high-throughput experiments, such as the ATLAS experiment at the CERN Large Hadron Collider \citep{ATLAS_ATLAS_2008}, data is generated at such a high velocity that real-time decisions must be made whether to save or discard individual samples. For this type of initial screening task, we imagine our low-latency and flexible prediction method will be an attractive candidate. 
Another experimental use-case for low-latency models is as replacements for force fields in molecular dynamics simulations \citep{gilmer_neural_nodate}. In these simulations, the energy and forces in a molecular system is repeatedly queried in a serial fashion. Any improvement in the latency of energy or force prediction would translate to a speedup of the overall system. Because our model has an extremely shallow computational graph, we expect it will have much lower latency than deep neural networks once implemented on a GPU.

One can also interpret our work as an initial investigation into the use of random Fourier feature methods for approximating common kernel methods in computational chemistry. 
Kernel methods are well-studied and highly performant in machine learning for computational chemistry; examples of such methods include \citet{rupp_fast_2012, montavon_learning_nodate, bartok_representing_2013, christensen_fchl_2020}. 
It is a natural question to ask whether these methods can benefit from low prediction latency while maintaining high test accuracy when approximated by random features. Our method does not implement an exact approximation to any of the kernel methods above, but our experiments show promise in this area. Our test errors are near those of FCHL19 on the QM7 dataset, and versions of our model using only 10\% of the optimal number of parameters have reasonable test errors. However, when training our model on the QM9 dataset, we have seen that the particular element-type encoding we have used creates ill-conditioned feature matrices and makes optimization difficult. To fully realize the potential of random feature methods in accelerating kernel learning for computational chemistry, a new element-type encoding is needed. 

%% file: broader_impacts.tex
\section{Broader Impact Statement}
The task of molecular property prediction is important to a wide range of applications, including the development of new pharmaceuticals, materials, and solvents. Some of these applications may be misused.

%% file: acknowledgements.tex
\section*{Acknowledgments}
RMW was supported by AFSOR award FA9550-18-1-0166 and NSF DMS-2023109.
OJM was supported by an NSF Research Traineeship under grant NSF 2022023.

%% file: appendix_integration_details.tex
\section{Full Integration Details} \label{sec:full_integration_details}
We wish to solve the following integral, which appears in the definition of rotation-invariant random features \cref{eq:invariant_random_features_integral}:
\begin{equation}
  \int_{SO(3)} \langle Q \circ p_i, g \rangle^2 dQ
\end{equation}
From  \cref{sec:evaluating_random_features}, we know this integral can decompose into
\begin{equation}
  \int_{SO(3)} \langle Q \circ p_i, g \rangle^2 dQ = \sum_{j_1, j_2 = 1}^{N_i} \int_{SO(3)} g(Qx_{i,{j_1}})g(Qx_{i,{j_2}}) dQ
\end{equation}
And we also have chosen a particular functional form for our random function $g$:
\begin{equation}
  g(x) = \sum_{k=1}^K \sum_{\ell = 0}^L \sum_{m=- \ell}^\ell w^{(\ell)}_{m,k} Y^{(\ell)}_m(\hat{x}) R_k(\| x \|)
\end{equation}
Repeated application of the linearity of integration gives us:
\begin{align}
  &\int_{SO(3)} \langle Q \circ p_i, g \rangle^2 dQ = \\
  =& \sum_{j_1, j_2 = 1}^N \int_{SO(3)} g(Qx_{i,{j_1}})g(Qx_{i,{j_2}}) dQ
\end{align}
\begin{multline}
  = \sum_{j_1, j_2 = 1}^{N_i} \sum_{k_1,k_2 = 1}^K \sum_{\ell_1, \ell_2 = 0}^L \sum_{m_1 = - \ell_1}^{\ell_1}\sum_{m_2 = - \ell_2}^{\ell_2} w^{(\ell_1)}_{m_1,k_1} w^{(\ell_2)}_{m_2,k_2} R_{k_1}(\| x_{i,{j_1}} \|) R_{k_2}(\| x_{i,{j_2}} \|) 
  \\ \: \: 
\int_{SO(3)} Y^{(\ell_1)}_{m_1}(Q \hat{x}_{i,{j_1}})Y^{(\ell_2)}_{m_2}(Q \hat{x}_{i,{j_2}}) dQ
\label{eq:expanded_random_feature_expression_1}
\end{multline}
We can now focus on the integral in  \cref{eq:expanded_random_feature_expression_1}, apply the rotation rule for spherical harmonics introduced in  \cref{sec:spherical_harmonics}, and again exploit the linearity of the integral.
\begin{align}
  &\int_{SO(3)} Y^{(\ell_1)}_{m_1}(Q \hat{x}_{i,{j_1}})Y^{(\ell_2)}_{m_2}(Q \hat{x}_{i,{j_2}}) dQ \\
  &= \int_{SO(3)} \left( \sum_{m'_1 = -\ell_1}^{\ell_1} Y_{m'_1}^{(\ell_1)}(\hat{x}_{i,{j_1}}) D^{(\ell_1)}(Q)_{m_1,m'_1} \right) \left(
  \sum_{m'_2 = -\ell_2}^{\ell_2} Y_{m'_2}^{(\ell_2)}(\hat{x}_{i,{j_2}}) D^{(\ell_2)}(Q)_{m_2,m'_2} \right) dQ \\
  &= \sum_{m'_1 = -\ell_1}^{\ell_1} \sum_{m'_2 = -\ell_2}^{\ell_2} Y_{m'_1}^{(\ell_1)}(\hat{x}_{i,{j_1}})  Y_{m'_2}^{(\ell_2)}(\hat{x}_{i,{j_2}}) 
  \int_{SO(3)}  D^{(\ell_1)}(Q)_{m_1,m'_1} D^{(\ell_2)}(Q)_{m_2,m'_2} dQ \\
\end{align}
At this point, we are able to apply the integration rule for elements of Wigner-D matrices introduced in \cref{sec:spherical_harmonics}.
\begin{align}
  &\int_{SO(3)} Y^{(\ell_1)}_{m_1}(Q \hat{x}_{i,{j_1}})Y^{(\ell_2)}_{m_2}(Q \hat{x}_{i,{j_2}}) dQ \\
  &= \sum_{m'_1 = -\ell_1}^{\ell_1} \sum_{m'_2 = -\ell_2}^{\ell_2} Y_{m'_1}^{(\ell_1)}(\hat{x}_{i,{j_1}})  Y_{m'_2}^{(\ell_2)}(\hat{x}_{i,{j_2}}) 
  (-1)^{m_1-m_1'} \frac{8 \pi^2}{2 \ell + 1} \delta_{-m_1, m_2} \delta_{-m_1', m_2'} \delta_{\ell_1, \ell_2} \\
  &= \delta_{\ell_1, \ell_2} \delta_{-m_1, m_2} \frac{8 \pi^2}{2 \ell + 1} \sum_{m_1' = - \ell_1}^{\ell_1} (-1)^{m_1-m_1'} Y_{m'_1}^{(\ell_1)}(\hat{x}_{i,{j_1}})Y_{-m'_1}^{(\ell_1)}(\hat{x}_{i,{j_2}})
  \label{eq:spherical_harmonic_product_integral}
\end{align}
To put it all together, we substitute  \cref{eq:spherical_harmonic_product_integral} into  \cref{eq:expanded_random_feature_expression_1}, and we see many terms drop out of the sums: 
\begin{multline}
  (\ref{eq:expanded_random_feature_expression_1}) =\sum_{j_1, j_2 = 1}^{N_i} \sum_{k_1,k_2 = 0}^K \sum_{\ell_1, \ell_2 = 0}^L \sum_{m_1 = - \ell_1}^{\ell_1}\sum_{m_2 = - \ell_2}^{\ell_2} 
  w^{(\ell_1)}_{m_1,k_1} w^{(\ell_2)}_{m_2,k_2} R_{k_1}(\| x_{i,{j_1}} \|) R_{k_2}(\| x_{i,{j_2}} \|) \\
  \delta_{\ell_1, \ell_2} \delta_{-m_1, m_2} \frac{8 \pi^2}{2 \ell + 1} \sum_{m_1' = - \ell_1}^{\ell_1} (-1)^{m_1-m_1'} Y_{m'_1}^{(\ell_1)}(\hat{x}_{i,{j_1}})Y_{-m'_1}^{(\ell_1)}(\hat{x}_{i,{j_2}})
\end{multline}
\begin{multline}
  \int_{SO(3)} \langle Q \circ p_i, g \rangle^2 dQ =\sum_{j_1, j_2 = 1}^{N_i} \sum_{k_1,k_2 = 0}^K \sum_{\ell_1 = 0}^L \sum_{m_1 = - \ell_1}^{\ell_1} 
  w^{(\ell_1)}_{m_1, k_1} w^{(\ell_1)}_{-m_1, k_2} R_{k_1}(\| x_{i,{j_1}} \|) R_{k_2}(\| x_{i,{j_2}} \|) \\
  \frac{8 \pi^2}{2 \ell + 1} \sum_{m_1' = - \ell_1}^{\ell_1} (-1)^{m_1-m_1'} Y_{m'_1}^{(\ell_1)}(\hat{x}_{i,{j_1}})Y_{-m'_1}^{(\ell_1)}(\hat{x}_{i,{j_2}})
  \label{eq:final_expression}
\end{multline}
Once the spherical harmonics and radial functions are evaluated, this sum has $O(N^2 K^2 L^3)$ terms. 

\subsection{Basis Expansion} \label{sec:basis_expansion}
When implementing  \cref{eq:final_expression} to compute rotationally-invariant random features, one can see that a large amount of computational effort can be re-used between different random features evaluated on the same sample. A table of spherical harmonic evaluations for all the points $\{\hat{x}_{i,1}, \hdots \hat{x}_{i,{N_i}} \}$ can be pre-computed once and re-used. Also, by re-arranging the order of summation in  \cref{eq:final_expression}, we arrive at this expression:
\begin{multline}
  \int_{SO(3)} \langle Q \circ p_i, g \rangle^2 dQ =  \sum_{k_1,k_2 = 1}^K \sum_{\ell_1 = 0}^L \sum_{m_1 = - \ell_1}^{\ell_1} 
  w^{(\ell_1)}_{m_1, k_1} w^{(\ell_1)}_{-m_1, k_2} \\
  \sum_{j_1, j_2 = 1}^{N_i} R_{k_1}(\| x_{i,{j_1}} \|) R_{k_2}(\| x_{i,{j_2}} \|)
  \frac{8 \pi^2}{2 \ell + 1} \sum_{m_1' = - \ell_1}^{\ell_1} (-1)^{m_1-m_1'} Y_{m'_1}^{(\ell_1)}(\hat{x}_{i,{j_1}})Y_{-m'_1}^{(\ell_1)}(\hat{x}_{i,{j_2}})
  \label{eq:rearranged_final_expression}
\end{multline}
Pre-computing everything after the line break allows for the re-use of a large amount of computational work between different random feature evaluations. We can interpret this pre-computation as an expansion of our point cloud in a rotationally-invariant basis $B$, which depends on the choice of radial function.  \cref{sec:correspondance_with_ACE} relates this basis expansion to the Atomic Cluster Expansion method.
\begin{align}
  B[\ell, m, k_1, k_2] &= \sum_{j_1, j_2 = 1}^{N_i} R_{k_1}(\| x_{i,{j_1}} \|) R_{k_2}(\| x_{i,{j_2}} \|)
  \frac{8 \pi^2}{2 \ell + 1}
  \sum_{m' = - \ell}^{\ell} (-1)^{m-m'} Y_{m'}^{(\ell)}(\hat{x}_{i,{j_1}})Y_{-m'}^{(\ell)}(\hat{x}_{i,{j_2}})
  \label{eq:precomputation_tensor} \\
  &= \sum_{j_1, j_2 = 1}^{N_i} R_{k_1}(\| x_{i,{j_1}} \|) R_{k_2}(\| x_{i,{j_2}} \|) \int_{SO(3)} Y^{(\ell)}_m(Q\hat{x}_{i,{j_1}})Y^{(\ell)}_{-m}(Q\hat{x}_{i,{j_2}}) dQ
  \label{eq:meaning_of_precomputation_tensor}
\end{align}
In our implementation we precompute this $B$ tensor once for each sample, and then we perform a tensor contraction with the random weights: 
\begin{align}
  \int_{SO(3)} \langle Q \circ p_i, g \rangle^2 dQ = \sum_{k_1,k_2 = 1}^K \sum_{\ell_1 = 0}^L \sum_{m_1 = - \ell_1}^{\ell_1} 
  w^{(\ell_1)}_{m_1, k_1} w^{(\ell_1)}_{-m_1, k_2} B[\ell_1, m_1, k_1, k_2]
\end{align}

%% file: appendix_connection_with_ACE.tex
\section{Connection Between Our Method and Randomized ACE Methods} \label{sec:correspondance_with_ACE}
\subsection{Overview of the ACE basis}
The atomic cluster expansion (ACE) method builds a rotationally-invariant basis for point clouds that are an extension of our pre-computed basis expansion described in  \cref{sec:basis_expansion}. 
In the ACE method, a preliminary basis $A_{k, \ell, m}$ is formed by projecting a point cloud function $p_i(x) = \sum_{j} \delta(x - x_{i_j})$ onto a single-particle basis function $\varphi_{k, \ell, m}(x)$:
\begin{align}
    \varphi_{k, \ell, m}(x) &= R_k(\| x \| ) Y^{(\ell)}_m(x) \\
    A_{k, \ell, m}(p_i) &= \langle \varphi_{k, \ell, m},  p_i \rangle \\
    &= \sum_{j} \varphi_{k, \ell, m}(x_{i_j})
\end{align}
where $R_k$ is a set of radial functions. In \cite{kovacs_linear_2021}, this set of radial functions is defined by taking a nonlinear radial transformation, and then the radial functions are a set of orthogonal polynomials defined on the transformed radii. This basis set is augmented with auxiliary basis functions to ensure the potential energy of two atoms at extremely nearby points diverges to infinity. 

The $A_{k, \ell, m}$ basis is a first step, but it only models single particles at a time. To model pairwise particle interactions, or higher-order interactions, the $A$ basis is extended by taking tensor products. First, a ``correlation order'' $N$ is chosen. Then, the $A$ basis is extended via tensor products: 
\begin{align}
    A_{\underline{k}, \underline{\ell}, \underline{m}}(p_i) = \prod_{\alpha=1}^N A_{k_\alpha, \ell_\alpha, m_\alpha} (p_i)
\end{align}
where $\underline{k} = (k_\alpha)_{\alpha=1}^N$, $\underline{\ell} = (\ell_\alpha)_{\alpha=1}^N$, and $\underline{m} = (m_\alpha)_{\alpha=1}^N$ are multi-indices. 

Finally, to ensure invariance with respect to permutations, reflections, and rotations of the point cloud, the ACE method forms a Haar measure $dg$ over $O(3)$ and computes a symmetrized version of the $A$ basis: 
\begin{align}
    B_{\underline{k}, \underline{\ell}, \underline{m}}(p_i) = \int_{O(3)}  A_{\underline{k}, \underline{\ell}, \underline{m}}(H \circ p_i) dH
\end{align}
\citet{drautz_atomic_2019} gives a detailed description of how this integral is performed. Applications of representation theory give concise descriptions of which multi-indices $\underline{k}, \underline{\ell}, \underline{m}$ remain after performing this integral; many are identically zero. 
\subsection{Connection to Our Method}
If one chooses to compute an ACE basis with correlation order $N=2$, the resulting basis is very similar to our pre-computed basis expansion in  \cref{eq:meaning_of_precomputation_tensor}. 
\begin{align}
    B_{\underline{k}, \underline{\ell}, \underline{m}}(p_i) &= \int_{O(3)}  A_{\underline{k}, \underline{\ell}, \underline{m}}(H \circ p_i) dH \\
   &= \int_{O(3)}  A_{k_1, \ell_1, m_1}(H \circ p_i) A_{k_2, \ell_2, m_2}(H \circ p_i) dH \\
   &= \int_{O(3)} \left( \sum_{j_1} \varphi_{k_1, \ell_1, m_1}(H x_{i_{j_1}}) \right)\left( \sum_{j_2} \varphi_{k_2, \ell_2, m_2}(H x_{i_{j_2}}) \right) dH \\
   &= \int_{O(3)} \left( \sum_{j_1} R_{k_1}(\| x_{i_{j_1}} \|) Y^{(\ell_1)}_{m_1}(H \hat{x}_{i_{j_1}}) \right) \left( \sum_{j_2} R_{k_2}(\| x_{i_{j_2}} \|) Y^{(\ell_2)}_{m_2}( H \hat{x}_{i_{j_2}}) \right) dH \\
  &= \sum_{j_1, j_2} R_{k_1}(\| x_{i_{j_1}} \|) R_{k_2}(\| x_{i_{j_2}} \|) \int_{O(3)}  Y^{(\ell_1)}_{m_1}(H \hat{x}_{i_{j_1}})   Y^{(\ell_2)}_{m_2}( H \hat{x}_{i_{j_2}}) dH 
\end{align}
The remaining difference between our basis expansion in  \cref{eq:meaning_of_precomputation_tensor} and the ACE basis expansion with correlation order $N=2$ is the ACE basis integrates over all of $O(3)$, while our method only enforces invariance with respect to $SO(3) \subset O(3)$.

So one can describe our random features as random nonlinear projections of a simplified version of the ACE basis with correlation order $N=2$ using extremely simple radial functions. 

\subsection{Benefit of Random Features}
We conduct the following experiment to quantify the benefit of using random nonlinear features over fitting a linear model in our simplified version of the ACE basis. Using the same hyperparameters as in our energy regression experiments, we compute our basis expansion for all the samples in the QM7 dataset. Using the same train/validation/test split, we perform ridge regression by fitting a linear model in our basis $B$ with $L^2$ regularization. We use the validation set to identify the best-performing model and report errors on the test set in  \cref{tab:benefit_of_random_features}. We note that the best-performing linear model has more than twice the error of the best performing random features model. 
\begin{table}
    \centering
    \begin{tabular}{p{0.4\linewidth}r}\toprule
        \textbf{Method}  & \textbf{Test Error (eV)}  \\ \midrule 
        Spherical CNNs & 0.1565   \\
        FCHL19         &  0.0541  \\
        Random Features (Ours) & 0.0660  \\
        Linear Model of $B$ basis (Ours) & 0.1542 \\
        \bottomrule \\
    \end{tabular}
    \caption[Quantifying the benefit of random features]{Reported are test errors on the QM7 dataset for Spherical CNNs, FCHL19, and two variants of our method. The first variant is the full rotation-invariant random features model. The second variant is a linear model of our basis expansion. We note that while our random feature model is the second-best performing model on this benchmark, our linear model is the nearly the worst-performing model.}
    \label{tab:benefit_of_random_features}
\end{table}

%% file: appendix_representation_theory.tex
\section{Representation Theory of the 3D Rotation Group}
\label{sec:Wigner_D_matrix_proof}

\subsection{Conjugation of Wigner-D Matrices}
In the following, we derive a formula for conjugating an element of a Wigner-D matrix. First, we use definition 6.44 in \cite{thompson_angular_1994} for the Wigner-D matrices:
\begin{equation}
    D^{(\ell)}_{m', m}(\alpha, \beta, \gamma) = e^{-i m' \alpha} d^{(\ell)}_{m', m}(\beta) e^{-i m \gamma}
\end{equation}
Here, $(\alpha, \beta, \gamma)$ are Euler angles parameterizing a rotation, and $d^{(\ell)}_{m', m}$ is an element of a Wigner-d (small d) matrix. We omit the exact definition of $d^{(\ell)}_{m', m}$, but we note the following symmetry properties:
\begin{align}
    d^{(\ell)}_{-m', -m}(\beta) &= d^{(\ell)}_{m', m}(-\beta) \\ 
     d^{(\ell)}_{m', m}(-\beta) &= d^{(\ell)}_{m, m'}(\beta) \\
     d^{(\ell)}_{m, m'}(\beta) &= (-1)^{m-m'}d^{(\ell)}_{m', m}(\beta)
\end{align}
The relationship between the Wigner d matrix elements allows us to derive a formula for conjugating an element of a Wigner D matrix:
\begin{align}
    D^{(\ell)}_{m', m}(\alpha, \beta, \gamma)^* &= e^{i m' \alpha} d^{(\ell)}_{m', m}(\beta)^* e^{i m \gamma} \\
    &= e^{-i (-m') \alpha} d^{(\ell)}_{m', m}(\beta) e^{-i (-m) \gamma} \\
    &= e^{-i (-m') \alpha} d^{(\ell)}_{-m', -m}(-\beta) e^{-i (-m) \gamma} \\
    &= e^{-i (-m') \alpha} d^{(\ell)}_{-m, -m'}(\beta) e^{-i (-m) \gamma} \\   
    &= e^{-i (-m') \alpha}(-1)^{-m + m'} d^{(\ell)}_{-m', -m}(\beta) e^{-i (-m) \gamma} \\   
    &= (-1)^{m'-m}D^{(\ell)}_{-m', -m}(\alpha, \beta, \gamma)
\end{align}

\subsection{Orthogonality Relations}
We use the following orthogonality relationship between elements of Wigner D matrices:
\begin{align}
  \int_{SO(3)} D^{(\ell_1)}(Q)_{m_1,k_1}^* D^{(\ell_2)}(Q)_{m_2,k_2} dQ = \frac{8 \pi^2}{2 \ell + 1} \delta_{m_1, m_2} \delta_{k_1, k_2} \delta_{\ell_1, \ell_2}
\end{align}
This is a well-known fact about the Wigner-D matrices. For a proof, one can derive this relationship from Schur orthogonality relations, which are a consequence of Schur's lemma. A direct calculation can be found in section 6.4.2 of \cite{thompson_angular_1994}.
Combining the two facts from above, we arrive at the following identity, which we use to compute our rotation-invariant random features.

\begin{align}
  \int_{SO(3)} D^{(\ell_1)}(Q)_{m_1,k_1} D^{(\ell_2)}(Q)_{m_2,k_2} dQ = (-1)^{m_1-k_1} \frac{8 \pi^2}{2 \ell + 1} \delta_{-m_1, m_2} \delta_{-k_1, k_2} \delta_{\ell_1, \ell_2}
\end{align}

%% file: appendix_extra_latency_experiments.tex
\section{Additional Latency Experiments}
\label{sec:extra_latency_experiments}
\begin{figure}
    \begin{subfigure}[b]{0.5\textwidth}
        \centering
        \includegraphics[width=\textwidth, trim={0 0 15cm 0}, clip]{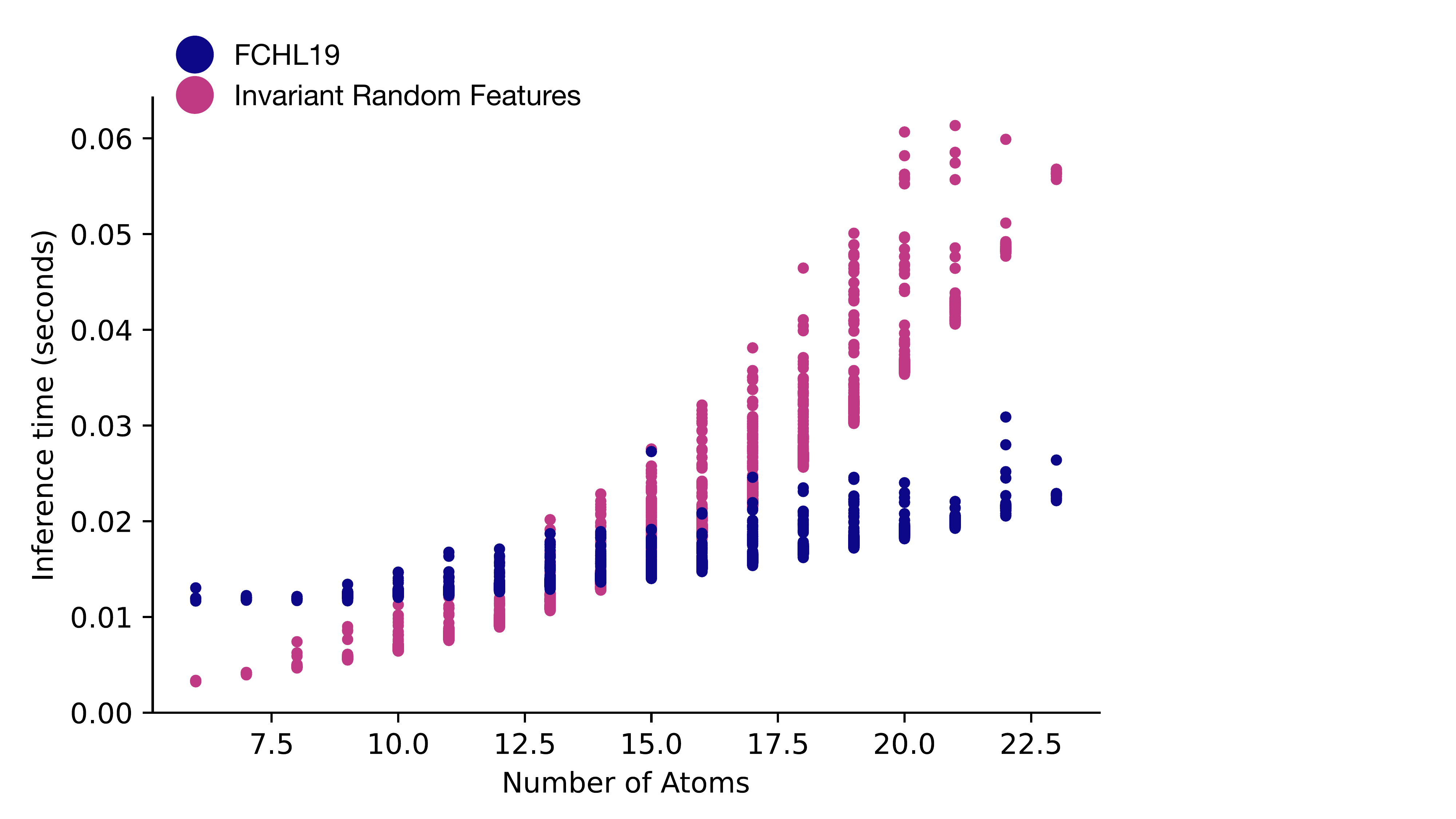}
        \caption{Latency for small models}
        \label{fig:small_models_latency}
    \end{subfigure}
    \begin{subfigure}[b]{0.5\textwidth}
        \centering
        \includegraphics[width=\textwidth, trim={0 0 15cm 0}, clip]{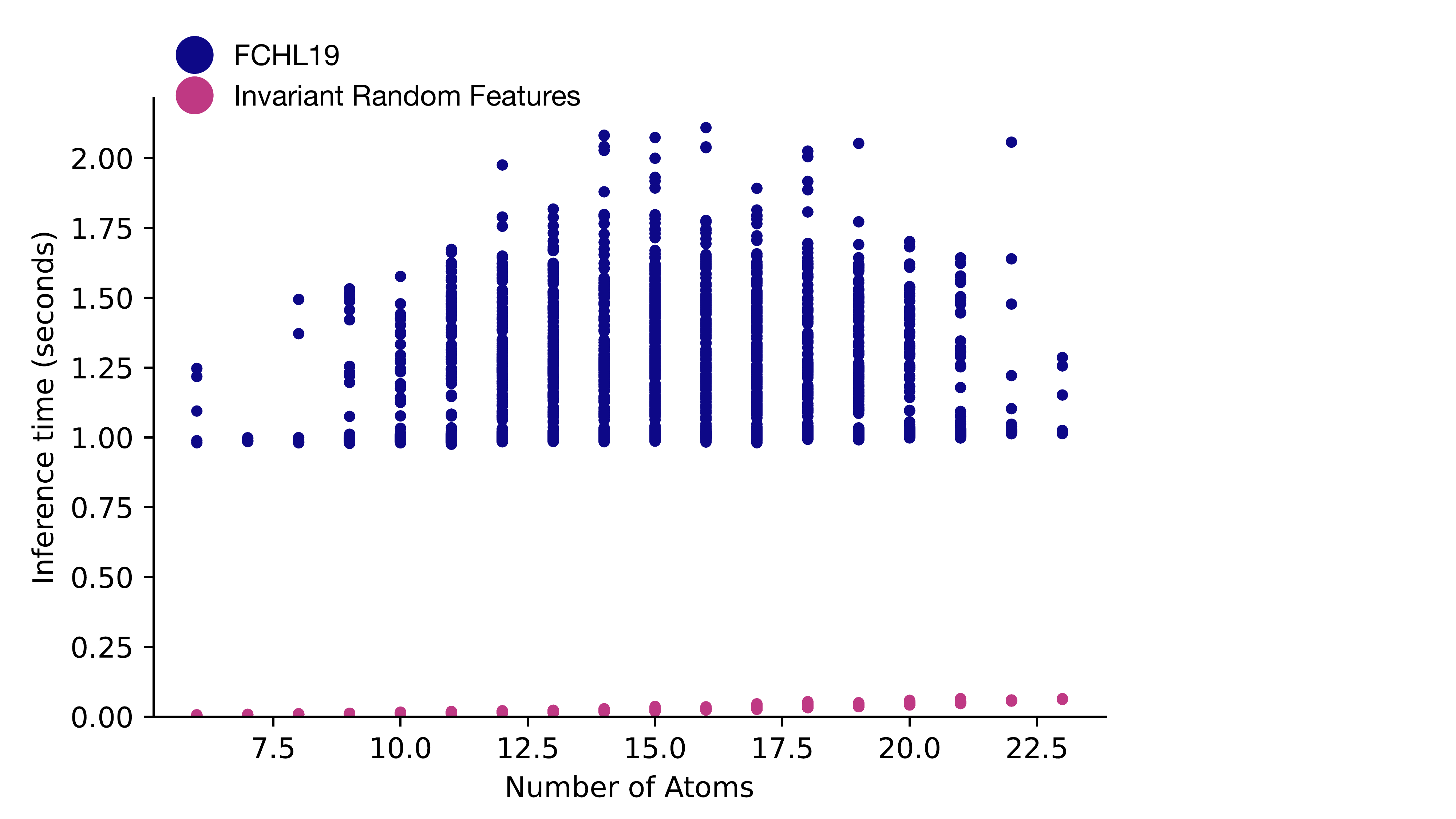}
        \caption{Latency for large models}
        \label{fig:large_models_latency}
    \end{subfigure}
    \caption[Prediction latency as a function of number of atoms.]{The number of atoms in a molecule affects the prediction latency for both FCHL19 and the invariant random features method. In \cref{fig:small_models_latency}, we show the prediction latencies for the smallest models tested, corresponding to 250 random features for our method and 200 training samples for FCHL19. For this model size, a quadratic scaling in the number of atoms largely determines the prediction latency. \cref{fig:large_models_latency} shows that for larger models (2,000 random features for our method and 5,000 training samples for FCHL19), the random features method is still in a regime where prediction latency is determined by the number of atoms, while FCHL19's prediction latency is dominated by the number of inner products required to evaluate the kernel.}
    \label{fig:latency_by_n_atoms}
\end{figure}

\begin{figure}
    \centering
    \includegraphics[width=0.7\textwidth]{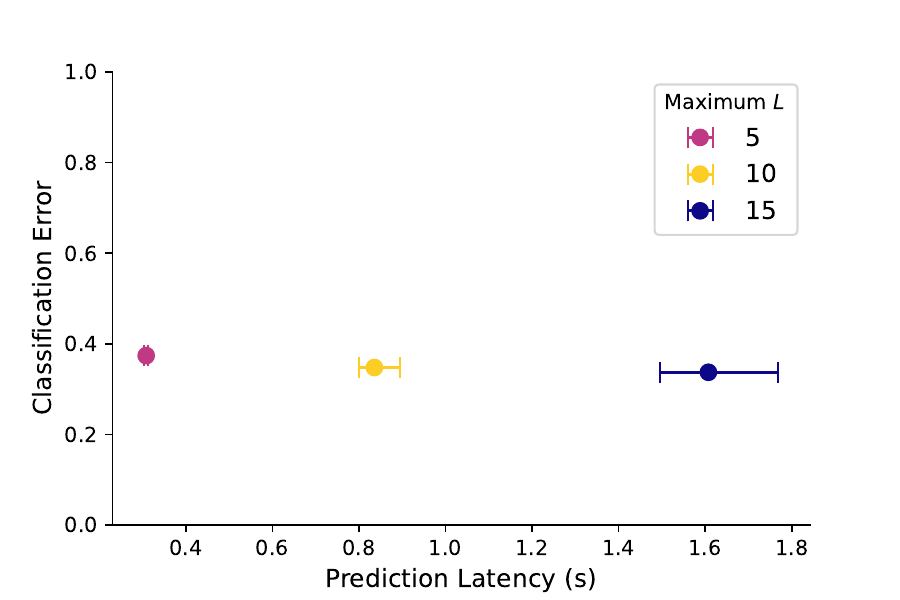}
    \caption[Prediction Latency on ModelNet40]{When training classification models on the ModelNet40 dataset, we are able to trade off prediction accuracy for prediction latency by changing the maximum frequency of spherical harmonics $L$. We represent samples from the ModelNet40 dataset using point clouds of size $N$=1,024. Computing random features has computational complexity scaling quadratically in $N$, which significantly slows down prediction latency. However, the prediction error of our model is relatively insensitive to the maximum frequency $L$ (see \cref{fig:hyperparams_modelnet40_max_L}), so we can greatly reduce the prediction latency by decreasing $L$. We evaluate prediction latency on a machine with 24 physical CPU cores. In the figure, the centers show the median latency sample, and the error bars span the $25^{th}$ and $75^{th}$ percentiles.}
    \label{fig:modelnet_prediction_latency}
\end{figure}

In this section we present extra experiments about the prediction latency of our method. \cref{fig:latency_by_n_atoms} compares the effect of molecule size on prediction latency for our method and FCHL19. Importantly, both methods have a precomputation step that has complexity $O(N^2)$ where $N$ is the number of atoms. For our experiments on ModelNet40, the quadratic dependence on the number of points causes slow prediction latency, but we are able to alleviate this latency by reducing other hyperparameters (\cref{fig:modelnet_prediction_latency}).  

\begin{figure}
    \centering
    \includegraphics[width=0.7\textwidth]{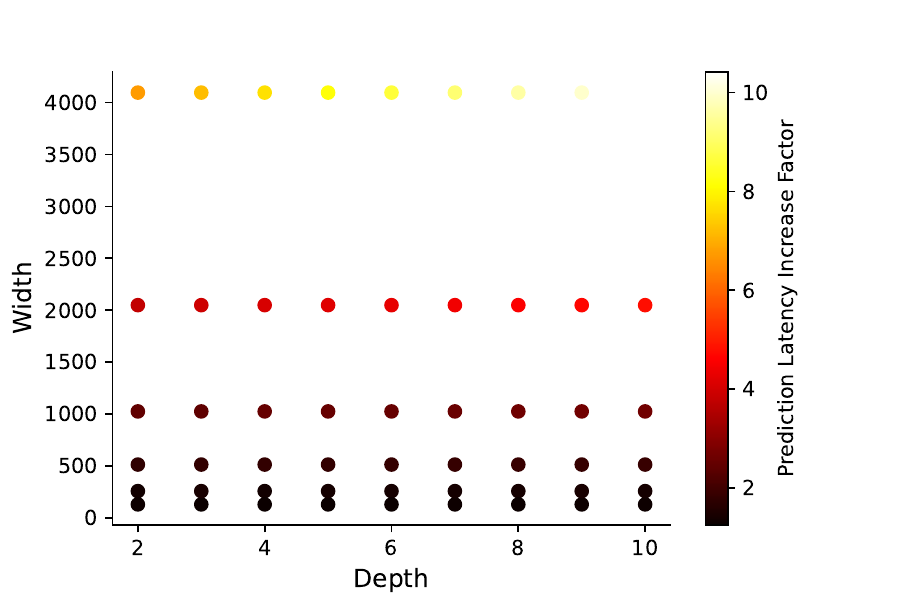}
    \caption[MLP prediction latencies]{We measure the prediction latency of computing 2,000 random features and predicting using different MLP architectures. We present increase in latency when compared to ridge regression. A value of 2.0 indicates that predictions take twice as long as predictions using ridge regression.}
    \label{fig:mlp_prediction_latencies}
\end{figure}

\begin{figure}[h!]
    \begin{subfigure}[b]{0.5\textwidth}
        \centering
        \includegraphics[width=\textwidth]{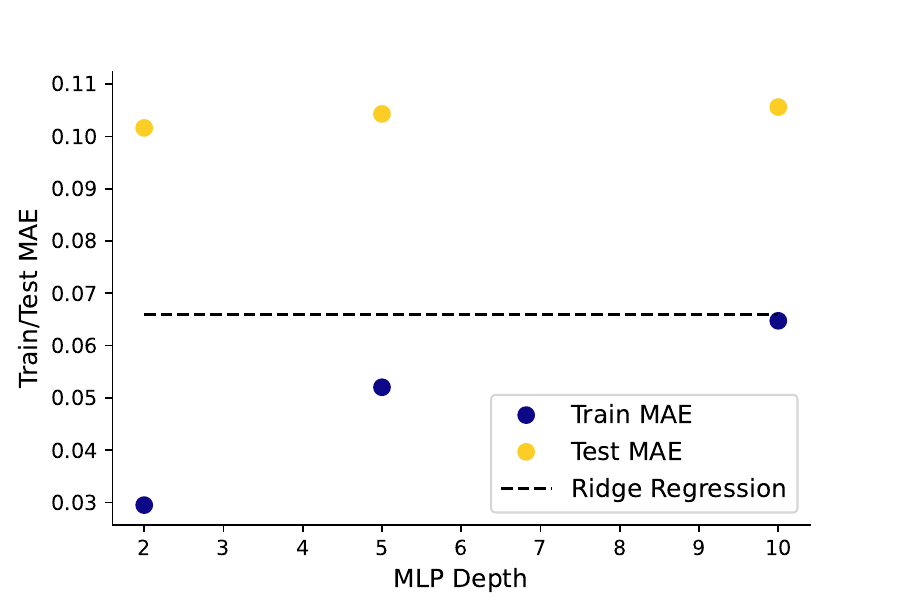}
        \caption{MLPs with width $=512$}
        \label{fig:mlp_error_512}
    \end{subfigure}
    \begin{subfigure}[b]{0.5\textwidth}
        \centering
        \includegraphics[width=\textwidth]{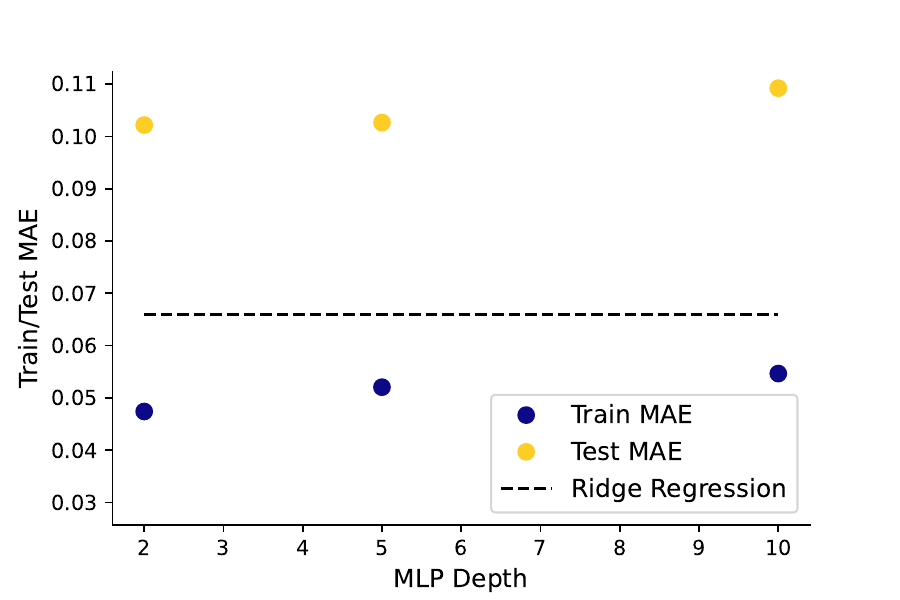}
        \caption{MLPs with width $=1024$}
        \label{fig:mlp_error_1024}
    \end{subfigure}
    \caption[Training Low-Latency MLPs.]{The ridge regression baseline outperforms MLP architectures with low prediction latency. }
    \label{fig:mlp_prediction_error}
\end{figure}

We also investigate whether using deep multilayer perceptrons (MLPs) provides an advantageous accuracy-latency tradeoff compared to our original ridge regression method. We first investigate the prediction latency incurred by using MLPs with different widths and depths. 
In \cref{fig:mlp_prediction_latencies}, we observe that models with widths 512 and 1024 only moderately increase prediction latency and are therefore good architecture candidates for the prediction step. Models with higher widths incur up to four times the prediction latency of ridge regression.
However, in \cref{fig:mlp_prediction_error}, we find that after training these architectures to convergence, their test error does not outperform ridge regression. 
The input to our MLPs were 2,000 random features from the QM7 dataset. We trained the MLPs to predict atomization energy by minimizing the mean squared error using the Adam optimizer. We searched over learning rate schedules and weight decay hyperparameters. 

%% file: appendix_extra_figures.tex
\section{Hyperparameter Sensitivity Analysis}
\label{sec:supplementary_figures}
\begin{figure}[h!]
    \begin{subfigure}[b]{0.5\textwidth}
        \centering
        \includegraphics[width=\textwidth]{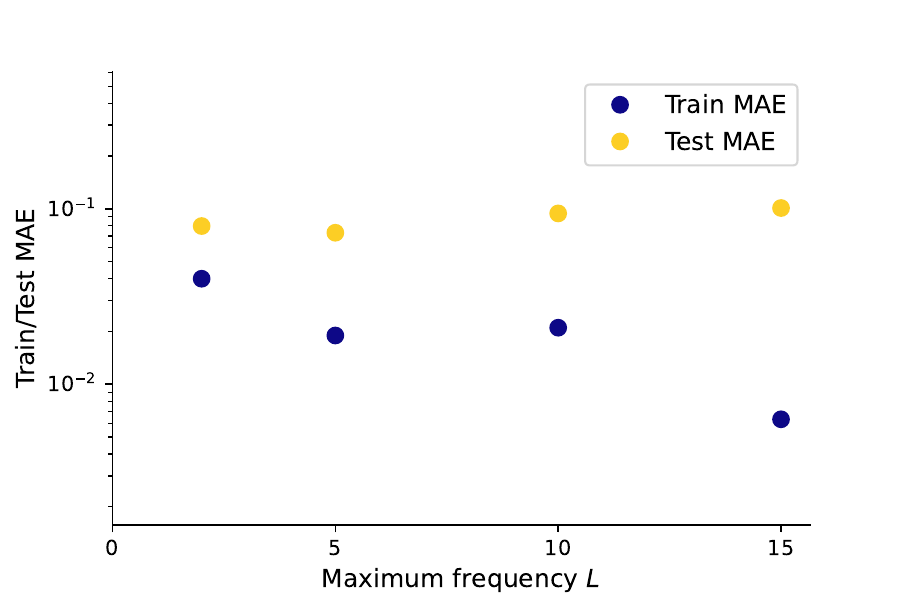}
        \caption{Effect of maximum frequency $L$}
        \label{fig:hyperparams_qm7_max_L}
    \end{subfigure}
    \begin{subfigure}[b]{0.5\textwidth}
        \centering
        \includegraphics[width=\textwidth]{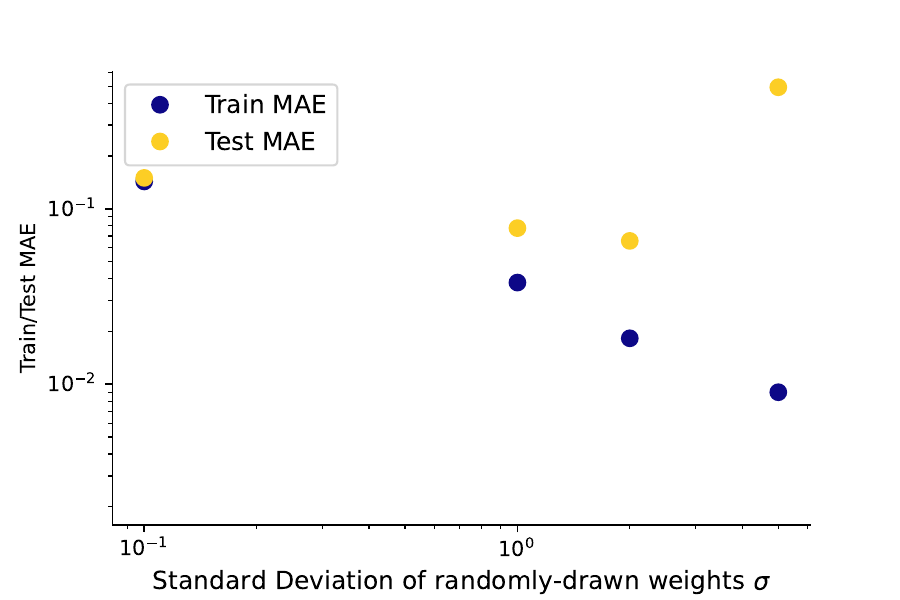}
        \caption{Effect of random weight distribution $\mathcal{N}(0, \sigma^2)$}
        \label{fig:hyperparams_qm7_weight_stddev}
    \end{subfigure}
    \caption[Hyperparameters for QM7 experiments.]{In the QM7 experiments, the standard deviation of the weights is the most important hyperparameter. For this sensitivity experiment, we use a standard hyperparameter setting and vary one hyperparameter at a time. The standard setting is 2,000 random features; standard deviation of the random weights $\sigma = 2.0$; and the maximum frequency $L = 5$. In all experiments, we search over a pre-defined grid of $L^2$ regularization parameters, and select the best model using a held-out validation set. In \cref{fig:hyperparams_qm7_max_L}, we vary the maximum frequency of the spherical harmonics, and we see this does not have a large effect on the test error. In \cref{fig:hyperparams_qm7_weight_stddev}, we vary $\sigma$, and we see this has a large effect on both the train and test error. }
    \label{fig:hyperparams_qm7}
\end{figure}

\begin{figure}[h!]
    \begin{subfigure}[b]{0.5\textwidth}
        \centering
        \includegraphics[width=\textwidth]{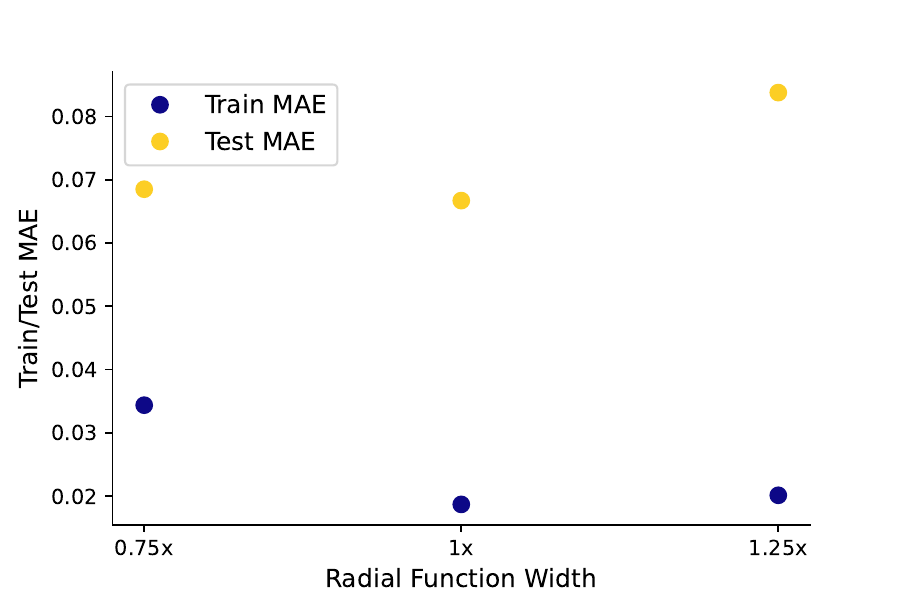}
        \caption{Effect of radial function width}
        \label{fig:radial_funcs_qm7_a}
    \end{subfigure}
    \begin{subfigure}[b]{0.5\textwidth}
        \centering
        \includegraphics[width=\textwidth]{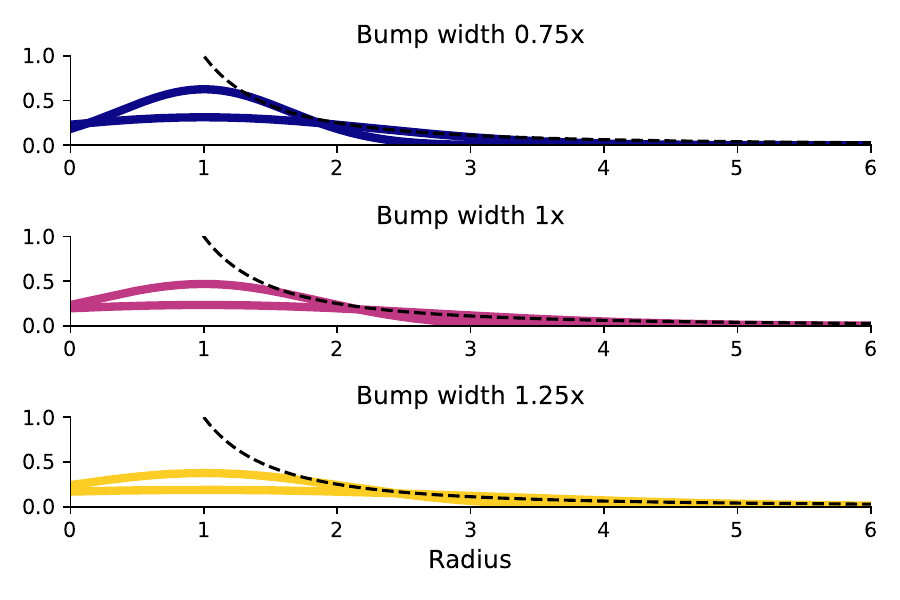}
        \caption{Visualizing different radial function widths}
        \label{fig:radial_funcs_qm7_b}
    \end{subfigure}
    \caption[Radial Functions for the QM7 experiments]{Our results on the QM7 dataset are slightly sensitive to the width of the chosen radial functions. For the molecular energy regression experiments, our radial functions are two Gaussians centered at $1$, with different widths $[\sigma_1, \sigma_2]$. We train and test models with different width scales: $0.75 \times [\sigma_1, \sigma_2]$ and $1.25 \times [\sigma_1, \sigma_2]$. We show the resulting train and test errors in \cref{fig:radial_funcs_qm7_a} and visualize the resulting radial functions in \cref{fig:radial_funcs_qm7_b}.}
    \label{fig:radial_funcs_qm7}
\end{figure}

\begin{figure}[h!]
    \begin{subfigure}[b]{0.5\textwidth}
        \centering
        \includegraphics[width=\textwidth]{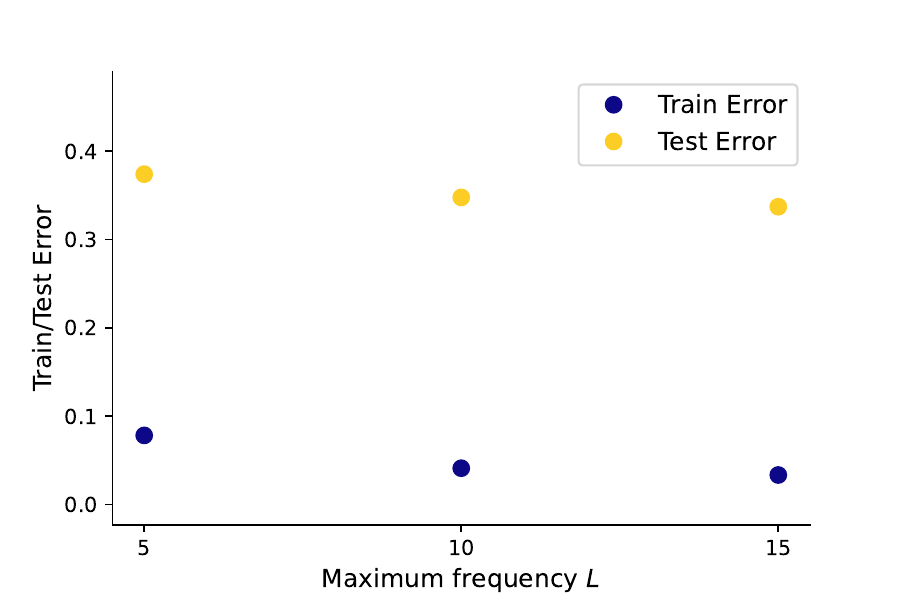}
        \caption{Effect of maximum frequency $L$}
        \label{fig:hyperparams_modelnet40_max_L}
    \end{subfigure}
    \begin{subfigure}[b]{0.5\textwidth}
        \centering
        \includegraphics[width=\textwidth]{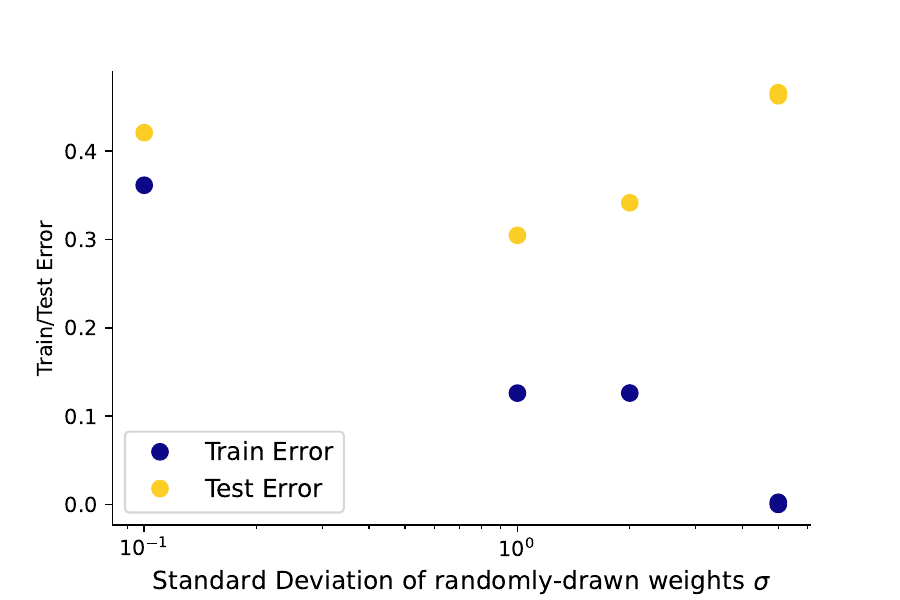}
        \caption{Effect of random weight distribution $\mathcal{N}(0, \sigma^2)$}
        \label{fig:hyperparams_modelnet40_weight_stddev}
    \end{subfigure}
    \caption[Hyperparameters for ModelNet40 experiments.]{In the ModelNet40 experiments, the standard deviation of the weights is the most important hyperparameter. For this sensitivity experiment, we use a standard hyperparameter setting and vary one hyperparameter at a time. The standard setting is 15,000 random features; standard deviation of the random weights $\sigma = 2.0$; and the maximum frequency $L = 10$. In \cref{fig:hyperparams_modelnet40_max_L}, we vary the maximum frequency of spherical harmonics $L$, and we see that increasing $L$ slightly improves the performance. In \cref{fig:hyperparams_modelnet40_weight_stddev}, we vary $\sigma$, and we see the model's performance is sensitive to this parameter.}
    \label{fig:hyperparams_modelnet40}
\end{figure}

\begin{figure}[h!]
    \begin{subfigure}[b]{0.5\textwidth}
        \centering
        \includegraphics[width=\textwidth]{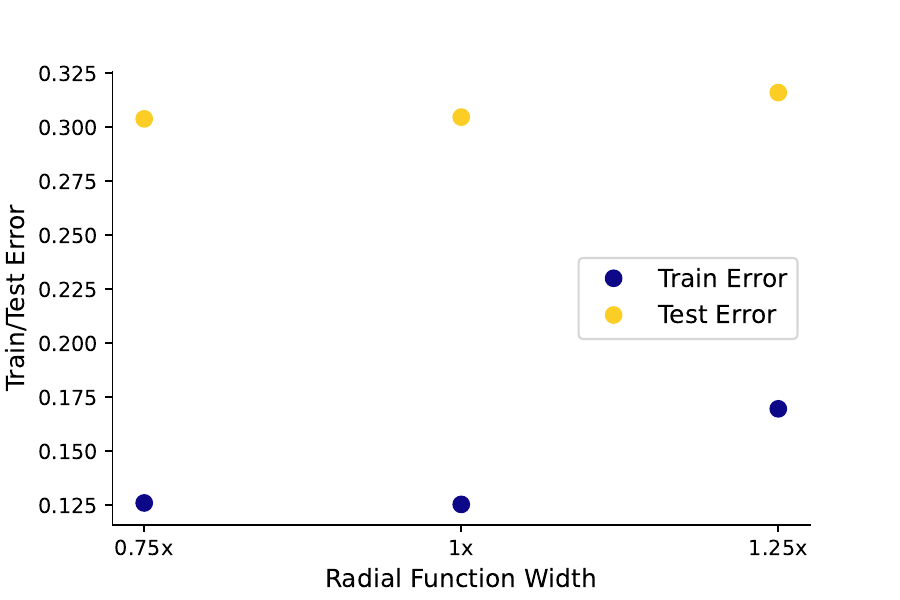}
        \caption{Effect of radial function width}
        \label{fig:radial_funcs_modelnet40_a}
    \end{subfigure}
    \begin{subfigure}[b]{0.5\textwidth}
        \centering
        \includegraphics[width=\textwidth]{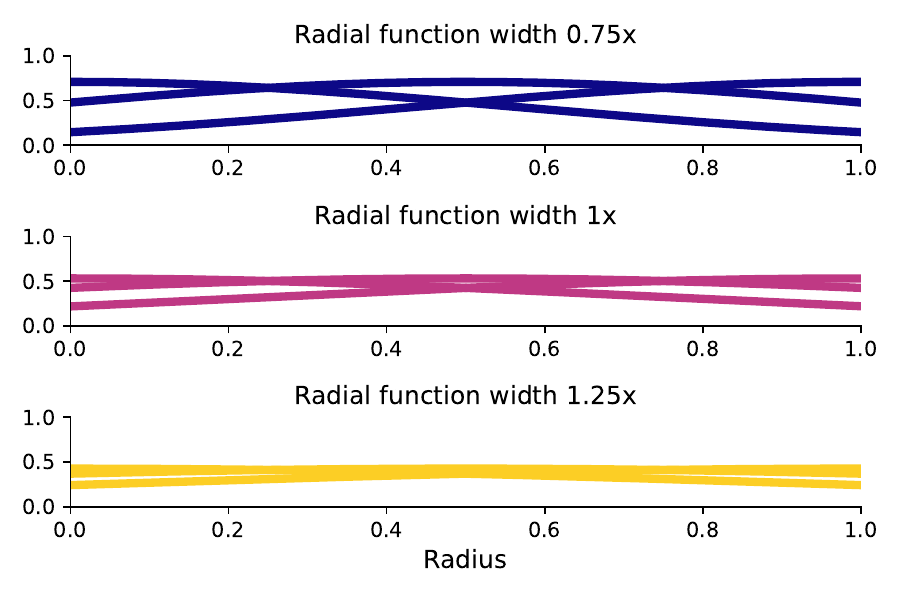}
        \caption{Visualizing different radial function widths}
        \label{fig:radial_funcs_modelnet40_b}
    \end{subfigure}
    \caption[Radial Functions for the ModelNet40 experiments]{Our results on the ModelNet40 dataset are very slightly sensitive to the width of the chosen radial functions. Our original radial functions are three Gaussians with centers $[0, 0.5, 1.0]$ and width $\sigma$. We train and test models with different radial function widths: $0.75 \times \sigma$, $1 \times \sigma$, and $1.25 \times \sigma$. \cref{fig:radial_funcs_modelnet40_a} shows the resulting train and test errors, and \cref{fig:radial_funcs_modelnet40_b} shows the resulting radial functions. }
    \label{fig:radial_funcs_modelnet40}
\end{figure}

We perform multiple experiments to evaluate the sensitivity of our random feature models' performance. In \cref{fig:hyperparams_qm7,fig:radial_funcs_qm7}, we find that our models trained on the QM7 dataset are most sensitive to $\sigma$, the standard deviation of the random weights. In \cref{fig:hyperparams_modelnet40,fig:radial_funcs_modelnet40}, we find the same result for models trained on the ModelNet40 dataset.

\clearpage

%% file: appendix_solving_least_squares_problems.tex
\section{Solving Ridge Regression Problems for Random Features}
\label{sec:ridge_regression_appendix}
This section outlines numerical methods for solving ridge regression problems arising in our random features models. 
In general, we want to solve the following problem:
\begin{equation}
    \argmin_\beta \| \Phi \beta - y \|_2^2 + \lambda \| \beta \|_2^2
    \label{eq:least_squares_problem_appendix}
\end{equation}
In our problem instances, $\Phi \in \R^{n \times d}$ is the matrix of random features. $\Phi$ is a dense matrix; there is no structured sparsity. $n$ is fixed; it is the number of samples in the dataset. The number of random features $d$ can be treated as a hyperparameter, but we find that $d > n$ is required for optimal test error, so the system of equations is underdetermined. $\lambda$ is the regularization parameter. Larger $\lambda$ corresponds to more regularization. 
Generally, when solving a large system, we approximately know a good value for $\lambda$ and want to keep it fixed.\footnote{As opposed to other optimization problems, where increasing $\lambda$ is an acceptable method of improving the conditioning of the system.}
To recap, we have an instance of a dense, overdetermined ridge regression problem.

By taking the gradient of \cref{eq:least_squares_problem_appendix} with respect to $\beta$ we arrive at the normal equation:
\begin{equation}
    0 = \left( \Phi^\top \Phi + \lambda I \right) \beta^* -  \Phi^\top y
    \label{eq:normal_equation_appendix}
\end{equation}

\subsection{Exact Solution via the Singular Value Decomposition}
From \cref{eq:normal_equation_appendix}, we see that the solution to the ridge regression problem is
\begin{align}
    \beta^* &= \left( \Phi^\top \Phi + \lambda I \right)^{-1}\Phi^\top y \nonumber \\
    &= V \left( \Sigma^2 + \lambda I \right)^{-1} \Sigma^\top U^\top y
    \label{eq:ridge_regression_solution}
\end{align}
where $\Phi = U \Sigma V^\top$ is the singular value decomposition (SVD). Because $\left( \Sigma^2 + \lambda I \right)$ is a diagonal matrix, we can exactly and efficiently compute its inverse. This suggests a method for solving \cref{eq:least_squares_problem_appendix}: compute the SVD of $\Phi$, efficiently compute $\left( \Sigma^2 + \lambda I \right)^{-1}$,  and reconstruct $\beta^*$ via \cref{eq:ridge_regression_solution}. This solution method is stable and relatively performant for small problem instances $(n, d < 10,000)$. We use this method in the QM7 experiments. Evaluating the full SVD of the feature matrix generated in the QM9 experiment is prohibitively costly, so we experimented with other solution methods.

\subsection{Approximate Solution via Iterative Methods}
\cref{eq:least_squares_problem_appendix} can be treated as an optimization problem in the variables $\beta \in \R^d$. The objective is strongly convex, with smoothness $\sigma^2_{\max}(\Phi)$ and strong convexity parameter\footnote{The strong convexity parameter is actually $\max \{\lambda, \sigma^2_{\min}(\Phi)\}$, but in practice, our random feature matrices are approximately low rank, so $\sigma^2_{\min}(\Phi) \ll \lambda$.} $\lambda$.  This motivates the application of standard iterative methods to find an approximate solution. Gradient descent, for instance, is known to enjoy linear convergence (on a log-log plot) for smooth and strongly-convex objectives. However, the rate of convergence is $\frac{\lambda}{\sigma^2_{\max}(\Phi)}$, which is very slow for typical problem instances. 

A family of iterative algorithms known as Krylov subspace methods are designed for linear least-squares problems like \cref{eq:least_squares_problem_appendix}. These methods include the conjugate gradient method and variations. In the QM9 experiment, we use LSQR \citep{paige_LSQR_1982,paige_algorithm-583_1982} implemented in scipy \citep{scipy}. 

While they are empirically faster than applying gradient descent to the objective and are known to converge to the exact solution in a finite number of iterations, the conjugate gradient method and LSQR both suffer from poor convergence dependence on the quantity $\frac{\sigma^2_{\max}(\Phi)}{\lambda}$. We have found that on our problem instance, LSQR requires approximately 100,000 iterations to converge to a solution with low training error. 

\subsection{Approximate Solution via Principal Components Regression}
We also attempt to approximately solve the ridge regression problem with principal components regression.
We compute a truncated SVD with scipy's sparse linear algebra package \citep{scipy}. The runtime of this operation depends on $k$, the number of singular vectors we choose to resolve. After computing the truncated SVD, we construct an approximate solution:
\begin{equation*}
    \hat{\beta}_k = V_k \left( \Sigma_k^2 + \lambda I \right)^{-1} \Sigma_k^\top U_k^\top y
\end{equation*}
Empirically, we find that we need to choose a large $k$ to find a solution with low training error, which incurs a large runtime. We find LSQR arrives at an approximate solution faster than principal components regression. 

\subsection{Future Work}
We have identified two different classes of methods from numerical linear algebra as promising candidates for improving our scaling to larger datasets. 

\subsubsection{Preconditioning Iterative Methods}
The runtime of iterative methods like LSQR \citep{paige_LSQR_1982,paige_algorithm-583_1982} depend adversely on the problem's conditioning $\frac{\sigma_{\max}^2(\Phi)}{\lambda}$. One way to improve the convergence of methods like LSQR is to find a preconditioning matrix $M \in \R^{n \times n}$ where $\sigma_{\max}^2(M\Phi) \ll \sigma_{\max}^2(\Phi)$. Once a preconditioner is formed, one solves the augmented problem 
\begin{equation*}
    \argmin_\beta \| M\Phi \beta - My \|_2^2 + \lambda \| \beta \|_2^2
\end{equation*}
using just a few iterations of an iterative solver. There are a number of classical preconditioning techniques and review articles, including \citet{wathen_preconditioning_2015,benzi_preconditioning_2002}.

LSRN \citep{meng_lsrn_2014} offers a promising method of computing a preconditioner designed for dense, overdetermined ridge regression problems. The algorithm constructs a preconditioner by subsampling columns of the matrix $\Phi$ and computing a singular value decomposition of the subsampled matrix. We believe that using a preconditioner like LSRN is a promising future research direction.

\subsubsection{Approximate Solution via Sketching}
Matrix sketching methods from the field of randomized numerical linear algebra \citep{drineas_randnla_2016} aim to reduce the dimension of a linear system while approximately preserving the solution. This can be performed by randomly selecting rows of the (possibly preconditioned) matrix $\Phi$ and solving the smaller problem instance. Sketching algorithms specifically for overdetermined ridge regression have been introduced in recent years \citet{chowdhury_iterative_2018,kacham_sketching_2022}.

%% file: paper.bbl
\begin{thebibliography}{47}
\providecommand{\natexlab}[1]{#1}
\providecommand{\url}[1]{\texttt{#1}}
\expandafter\ifx\csname urlstyle\endcsname\relax
  \providecommand{\doi}[1]{doi: #1}\else
  \providecommand{\doi}{doi: \begingroup \urlstyle{rm}\Url}\fi

\bibitem[Rahimi and Recht(2007)]{rahimi_random_2007}
Ali Rahimi and Benjamin Recht.
\newblock Random {Features} for {Large}-{Scale} {Kernel} {Machines}.
\newblock In J.~Platt, D.~Koller, Y.~Singer, and S.~Roweis, editors,
  \emph{Advances in {Neural} {Information} {Processing} {Systems}}, volume~20.
  Curran Associates, Inc., 2007.
\newblock URL
  \url{https://proceedings.neurips.cc/paper/2007/file/013a006f03dbc5392effeb8f18fda755-Paper.pdf}.

\bibitem[Kondor et~al.(2018)Kondor, Lin, and
  Trivedi]{kondor_clebsch-gordan_2018}
Risi Kondor, Zhen Lin, and Shubhendu Trivedi.
\newblock Clebsch–gordan nets: A fully fourier space spherical convolutional
  neural network.
\newblock In \emph{Proceedings of the 32nd International Conference on Neural
  Information Processing Systems}, NIPS'18, page 10138–10147, Red Hook, NY,
  USA, 2018. Curran Associates Inc.

\bibitem[Cohen et~al.(2018{\natexlab{a}})Cohen, Geiger, Köhler, and
  Welling]{cohen_spherical_2018}
Taco~S. Cohen, Mario Geiger, Jonas Köhler, and Max Welling.
\newblock Spherical {CNN}s.
\newblock In \emph{International Conference on Learning Representations},
  2018{\natexlab{a}}.
\newblock URL \url{https://openreview.net/forum?id=Hkbd5xZRb}.

\bibitem[Esteves et~al.(2018)Esteves, Allen-Blanchette, Makadia, and
  Daniilidis]{esteves_learning_2018}
Carlos Esteves, Christine Allen-Blanchette, Ameesh Makadia, and Kostas
  Daniilidis.
\newblock Learning {SO}(3) {Equivariant} {Representations} with {Spherical}
  {CNNs}, September 2018.
\newblock URL \url{http://arxiv.org/abs/1711.06721}.
\newblock arXiv:1711.06721 [cs].

\bibitem[Collaboration(2008)]{ATLAS_ATLAS_2008}
ATLAS Collaboration.
\newblock The atlas experiment at the cern large hadron collider.
\newblock \emph{Journal of Instrumentation}, 3\penalty0 (08):\penalty0 S08003,
  August 2008.
\newblock ISSN 1748-0221.
\newblock \doi{10.1088/1748-0221/3/08/S08003}.
\newblock URL \url{https://dx.doi.org/10.1088/1748-0221/3/08/S08003}.

\bibitem[Komiske et~al.(2018)Komiske, Metodiev, and
  Thaler]{komiske_energy_2018}
Patrick~T. Komiske, Eric~M. Metodiev, and Jesse Thaler.
\newblock Energy flow polynomials: A complete linear basis for jet
  substructure.
\newblock \emph{Journal of High Energy Physics}, 2018\penalty0 (4):\penalty0
  13, Apr 2018.
\newblock ISSN 1029-8479.
\newblock \doi{10.1007/JHEP04(2018)013}.
\newblock arXiv:1712.07124 [hep-ex, physics:hep-ph].

\bibitem[Thaler and Van~Tilburg(2011)]{thaler_identifying_2011}
Jesse Thaler and Ken Van~Tilburg.
\newblock Identifying boosted objects with n-subjettiness.
\newblock \emph{Journal of High Energy Physics}, 2011\penalty0 (3):\penalty0
  15, Mar 2011.
\newblock ISSN 1029-8479.
\newblock \doi{10.1007/JHEP03(2011)015}.

\bibitem[Bogatskiy et~al.(2022)Bogatskiy, Hoffman, Miller, and
  Offermann]{bogatskiy_pelican_2022}
Alexander Bogatskiy, Timothy Hoffman, David~W. Miller, and Jan~T. Offermann.
\newblock Pelican: Permutation equivariant and lorentz invariant or covariant
  aggregator network for particle physics.
\newblock \penalty0 (arXiv:2211.00454), Dec 2022.
\newblock URL \url{http://arxiv.org/abs/2211.00454}.
\newblock arXiv:2211.00454 [hep-ex, physics:hep-ph].

\bibitem[Gong et~al.(2022)Gong, Meng, Zhang, Qu, Li, Qian, Du, Ma, and
  Liu]{gong_an-efficient_2022}
Shiqi Gong, Qi~Meng, Jue Zhang, Huilin Qu, Congqiao Li, Sitian Qian, Weitao Du,
  Zhi-Ming Ma, and Tie-Yan Liu.
\newblock An efficient lorentz equivariant graph neural network for jet
  tagging.
\newblock \emph{Journal of High Energy Physics}, 2022\penalty0 (7):\penalty0
  30, Jul 2022.
\newblock ISSN 1029-8479.
\newblock \doi{10.1007/JHEP07(2022)030}.
\newblock arXiv:2201.08187 [hep-ex, physics:hep-ph].

\bibitem[Gilmer et~al.(2017)Gilmer, Schoenholz, Riley, Vinyals, and
  Dahl]{gilmer_neural_nodate}
Justin Gilmer, Samuel~S. Schoenholz, Patrick~F. Riley, Oriol Vinyals, and
  George~E. Dahl.
\newblock Neural message passing for quantum chemistry.
\newblock In \emph{Proceedings of the 34th International Conference on Machine
  Learning - Volume 70}, ICML'17, page 1263–1272. JMLR.org, 2017.

\bibitem[Christensen et~al.(2020)Christensen, Bratholm, and
  Faber]{christensen_fchl_2020}
Anders~S Christensen, Lars~A Bratholm, and Felix~A Faber.
\newblock {FCHL} revisited: {Faster} and more accurate quantum machine
  learning.
\newblock \emph{The Journal of Chemical Physics}, page~16, 2020.

\bibitem[Anderson et~al.(2019)Anderson, Hy, and
  Kondor]{anderson_cormorant_2019}
Brandon Anderson, Truong~Son Hy, and Risi Kondor.
\newblock Cormorant: Covariant molecular neural networks.
\newblock In H.~Wallach, H.~Larochelle, A.~Beygelzimer, F.~d\textquotesingle
  Alch\'{e}-Buc, E.~Fox, and R.~Garnett, editors, \emph{Advances in Neural
  Information Processing Systems}, volume~32. Curran Associates, Inc., 2019.
\newblock URL
  \url{https://proceedings.neurips.cc/paper/2019/file/03573b32b2746e6e8ca98b9123f2249b-Paper.pdf}.

\bibitem[Montavon et~al.(2012)Montavon, Hansen, Fazli, Rupp, Biegler, Ziehe,
  Tkatchenko, von Lilienfeld, and M\"{u}ller]{montavon_learning_nodate}
Gr\'{e}goire Montavon, Katja Hansen, Siamac Fazli, Matthias Rupp, Franziska
  Biegler, Andreas Ziehe, Alexandre Tkatchenko, O.~Anatole von Lilienfeld, and
  Klaus-Robert M\"{u}ller.
\newblock Learning invariant representations of molecules for atomization
  energy prediction.
\newblock In \emph{Proceedings of the 25th International Conference on Neural
  Information Processing Systems - Volume 1}, NIPS'12, page 440–448, Red
  Hook, NY, USA, 2012. Curran Associates Inc.

\bibitem[Rupp et~al.(2012)Rupp, Tkatchenko, Müller, and von
  Lilienfeld]{rupp_fast_2012}
Matthias Rupp, Alexandre Tkatchenko, Klaus-Robert Müller, and O.~Anatole von
  Lilienfeld.
\newblock Fast and {Accurate} {Modeling} of {Molecular} {Atomization}
  {Energies} with {Machine} {Learning}.
\newblock \emph{Physical Review Letters}, 108\penalty0 (5):\penalty0 058301,
  January 2012.
\newblock ISSN 0031-9007, 1079-7114.
\newblock \doi{10.1103/PhysRevLett.108.058301}.
\newblock URL \url{https://link.aps.org/doi/10.1103/PhysRevLett.108.058301}.

\bibitem[Bartók et~al.(2013)Bartók, Kondor, and
  Csányi]{bartok_representing_2013}
Albert~P. Bartók, Risi Kondor, and Gábor Csányi.
\newblock On representing chemical environments.
\newblock \emph{Physical Review B}, 87\penalty0 (18):\penalty0 184115, May
  2013.
\newblock ISSN 1098-0121, 1550-235X.
\newblock \doi{10.1103/PhysRevB.87.184115}.
\newblock URL \url{https://link.aps.org/doi/10.1103/PhysRevB.87.184115}.

\bibitem[Kovács et~al.(2021)Kovács, Oord, Kucera, Allen, Cole, Ortner, and
  Csányi]{kovacs_linear_2021}
Dávid~Péter Kovács, Cas van~der Oord, Jiri Kucera, Alice E.~A. Allen,
  Daniel~J. Cole, Christoph Ortner, and Gábor Csányi.
\newblock Linear {Atomic} {Cluster} {Expansion} {Force} {Fields} for {Organic}
  {Molecules}: {Beyond} {RMSE}.
\newblock \emph{Journal of Chemical Theory and Computation}, 17\penalty0
  (12):\penalty0 7696--7711, December 2021.
\newblock ISSN 1549-9618, 1549-9626.
\newblock \doi{10.1021/acs.jctc.1c00647}.
\newblock URL \url{https://pubs.acs.org/doi/10.1021/acs.jctc.1c00647}.

\bibitem[Drautz(2019)]{drautz_atomic_2019}
Ralf Drautz.
\newblock Atomic cluster expansion for accurate and transferable interatomic
  potentials.
\newblock \emph{Physical Review B}, 99\penalty0 (1):\penalty0 014104, January
  2019.
\newblock ISSN 2469-9950, 2469-9969.
\newblock \doi{10.1103/PhysRevB.99.014104}.
\newblock URL \url{https://link.aps.org/doi/10.1103/PhysRevB.99.014104}.

\bibitem[Schütt et~al.(2018)Schütt, Sauceda, Kindermans, Tkatchenko, and
  Müller]{schutt_schnet_2018}
K.~T. Schütt, H.~E. Sauceda, P.-J. Kindermans, A.~Tkatchenko, and K.-R.
  Müller.
\newblock {SchNet} – {A} deep learning architecture for molecules and
  materials.
\newblock \emph{The Journal of Chemical Physics}, 148\penalty0 (24):\penalty0
  241722, June 2018.
\newblock ISSN 0021-9606, 1089-7690.
\newblock \doi{10.1063/1.5019779}.
\newblock URL \url{http://aip.scitation.org/doi/10.1063/1.5019779}.

\bibitem[Unke and Meuwly(2019)]{unke_PhysNet_2019}
Oliver~T. Unke and Markus Meuwly.
\newblock {PhysNet}: {A} {Neural} {Network} for {Predicting} {Energies},
  {Forces}, {Dipole} {Moments}, and {Partial} {Charges}.
\newblock \emph{Journal of Chemical Theory and Computation}, 15\penalty0
  (6):\penalty0 3678--3693, June 2019.
\newblock ISSN 1549-9618, 1549-9626.
\newblock \doi{10.1021/acs.jctc.9b00181}.
\newblock URL \url{https://pubs.acs.org/doi/10.1021/acs.jctc.9b00181}.

\bibitem[Qiao et~al.(2020)Qiao, Welborn, Anandkumar, Manby, and
  Miller]{qiao_orbnet_2020}
Zhuoran Qiao, Matthew Welborn, Animashree Anandkumar, Frederick~R. Manby, and
  III Miller, Thomas~F.
\newblock Orbnet: Deep learning for quantum chemistry using symmetry-adapted
  atomic-orbital features.
\newblock \emph{The Journal of Chemical Physics}, 153\penalty0 (12):\penalty0
  124111, Sep 2020.
\newblock ISSN 0021-9606.
\newblock \doi{10.1063/5.0021955}.

\bibitem[Poulenard et~al.(2019)Poulenard, Rakotosaona, Ponty, and
  Ovsjanikov]{poulenard_effective_2019}
Adrien Poulenard, Marie-Julie Rakotosaona, Yann Ponty, and Maks Ovsjanikov.
\newblock Effective {Rotation}-{Invariant} {Point} {CNN} with {Spherical}
  {Harmonics} {Kernels}.
\newblock In \emph{2019 {International} {Conference} on {3D} {Vision} ({3DV})},
  pages 47--56, September 2019.
\newblock \doi{10.1109/3DV.2019.00015}.
\newblock ISSN: 2475-7888.

\bibitem[Thomas et~al.(2018)Thomas, Smidt, Kearnes, Yang, Li, Kohlhoff, and
  Riley]{thomas_tensor_2018}
Nathaniel Thomas, Tess Smidt, Steven Kearnes, Lusann Yang, Li~Li, Kai Kohlhoff,
  and Patrick Riley.
\newblock Tensor field networks: {Rotation}- and translation-equivariant neural
  networks for {3D} point clouds, May 2018.
\newblock URL \url{http://arxiv.org/abs/1802.08219}.
\newblock Number: arXiv:1802.08219 arXiv:1802.08219 [cs].

\bibitem[Guo et~al.(2021)Guo, Wang, Hu, Liu, Liu, and Bennamoun]{guo_deep_2021}
Yulan Guo, Hanyun Wang, Qingyong Hu, Hao Liu, Li~Liu, and Mohammed Bennamoun.
\newblock Deep {Learning} for {3D} {Point} {Clouds}: {A} {Survey}.
\newblock \emph{IEEE Transactions on Pattern Analysis and Machine
  Intelligence}, 43\penalty0 (12):\penalty0 4338--4364, December 2021.
\newblock ISSN 1939-3539.
\newblock \doi{10.1109/TPAMI.2020.3005434}.
\newblock Conference Name: IEEE Transactions on Pattern Analysis and Machine
  Intelligence.

\bibitem[Cohen et~al.(2018{\natexlab{b}})Cohen, Geiger, and
  Weiler]{cohen_intertwiners_2018}
Taco~S. Cohen, Mario Geiger, and Maurice Weiler.
\newblock Intertwiners between induced representations (with applications to
  the theory of equivariant neural networks), 2018{\natexlab{b}}.
\newblock URL \url{https://arxiv.org/abs/1803.10743}.

\bibitem[Weiler et~al.(2018)Weiler, Geiger, Welling, Boomsma, and
  Cohen]{weiler_3d_2018}
Maurice Weiler, Mario Geiger, Max Welling, Wouter Boomsma, and Taco Cohen.
\newblock 3d steerable cnns: Learning rotationally equivariant features in
  volumetric data.
\newblock In \emph{Proceedings of the 32nd International Conference on Neural
  Information Processing Systems}, NIPS'18, page 10402–10413, Red Hook, NY,
  USA, 2018. Curran Associates Inc.

\bibitem[Xiao et~al.(2020)Xiao, Lin, Li, Geng, Chao, and
  Ding]{xiao_endowing_2020}
Zelin Xiao, Hongxin Lin, Renjie Li, Lishuai Geng, Hongyang Chao, and Shengyong
  Ding.
\newblock Endowing {Deep} 3d {Models} {With} {Rotation} {Invariance} {Based}
  {On} {Principal} {Component} {Analysis}.
\newblock In \emph{2020 {IEEE} {International} {Conference} on {Multimedia} and
  {Expo} ({ICME})}, pages 1--6, London, United Kingdom, July 2020. IEEE.
\newblock ISBN 978-1-72811-331-9.
\newblock \doi{10.1109/ICME46284.2020.9102947}.
\newblock URL \url{https://ieeexplore.ieee.org/document/9102947/}.

\bibitem[Puny et~al.(2022)Puny, Atzmon, Smith, Misra, Grover, Ben-Hamu, and
  Lipman]{puny_frame_2022}
Omri Puny, Matan Atzmon, Edward~J. Smith, Ishan Misra, Aditya Grover, Heli
  Ben-Hamu, and Yaron Lipman.
\newblock Frame {Averaging} for {Invariant} and {Equivariant} {Network}
  {Design}.
\newblock In \emph{International {Conference} on {Learning} {Representations}},
  2022.
\newblock URL \url{https://openreview.net/forum?id=zIUyj55nXR}.

\bibitem[Deng et~al.(2021)Deng, Litany, Duan, Poulenard, Tagliasacchi, and
  Guibas]{deng_vector_2021}
Congyue Deng, Or~Litany, Yueqi Duan, Adrien Poulenard, Andrea Tagliasacchi, and
  Leonidas Guibas.
\newblock Vector {Neurons}: {A} {General} {Framework} for {SO}(3)-{Equivariant}
  {Networks}.
\newblock In \emph{2021 {IEEE}/{CVF} {International} {Conference} on {Computer}
  {Vision} ({ICCV})}, pages 12180--12189, Montreal, QC, Canada, October 2021.
  IEEE.
\newblock ISBN 978-1-66542-812-5.
\newblock \doi{10.1109/ICCV48922.2021.01198}.
\newblock URL \url{https://ieeexplore.ieee.org/document/9711441/}.

\bibitem[Villar et~al.(2021)Villar, Hogg, Storey-Fisher, Yao, and
  Blum-Smith]{villar_scalars_nodate}
Soledad Villar, David~W Hogg, Kate Storey-Fisher, Weichi Yao, and Ben
  Blum-Smith.
\newblock Scalars are universal: Equivariant machine learning, structured like
  classical physics.
\newblock In A.~Beygelzimer, Y.~Dauphin, P.~Liang, and J.~Wortman Vaughan,
  editors, \emph{Advances in Neural Information Processing Systems}, 2021.
\newblock URL \url{https://openreview.net/forum?id=ba27-RzNaIv}.

\bibitem[Rahimi and Recht(2008)]{rahimi_weighted_2008}
Ali Rahimi and Benjamin Recht.
\newblock Weighted {Sums} of {Random} {Kitchen} {Sinks}: {Replacing}
  minimization with randomization in learning.
\newblock In D.~Koller, D.~Schuurmans, Y.~Bengio, and L.~Bottou, editors,
  \emph{Advances in {Neural} {Information} {Processing} {Systems}}, volume~21.
  Curran Associates, Inc., 2008.
\newblock URL
  \url{https://proceedings.neurips.cc/paper/2008/file/0efe32849d230d7f53049ddc4a4b0c60-Paper.pdf}.

\bibitem[Mei et~al.(2021)Mei, Misiakiewicz, and Montanari]{mei_learning_2021}
Song Mei, Theodor Misiakiewicz, and Andrea Montanari.
\newblock Learning with invariances in random features and kernel models.
\newblock In Mikhail Belkin and Samory Kpotufe, editors, \emph{Proceedings of
  Thirty Fourth Conference on Learning Theory}, volume 134 of \emph{Proceedings
  of Machine Learning Research}, pages 3351--3418. PMLR, 15--19 Aug 2021.
\newblock URL \url{https://proceedings.mlr.press/v134/mei21a.html}.

\bibitem[Blum and Reymond(2009)]{blum_970_2009}
Lorenz~C. Blum and Jean-Louis Reymond.
\newblock 970 {Million} {Druglike} {Small} {Molecules} for {Virtual}
  {Screening} in the {Chemical} {Universe} {Database} {GDB}-13.
\newblock \emph{Journal of the American Chemical Society}, 131\penalty0
  (25):\penalty0 8732--8733, July 2009.
\newblock ISSN 0002-7863, 1520-5126.
\newblock \doi{10.1021/ja902302h}.
\newblock URL \url{https://pubs.acs.org/doi/10.1021/ja902302h}.

\bibitem[Ruddigkeit et~al.(2012)Ruddigkeit, van Deursen, Blum, and
  Reymond]{ruddigkeit_enumeration_2012}
Lars Ruddigkeit, Ruud van Deursen, Lorenz~C. Blum, and Jean-Louis Reymond.
\newblock Enumeration of 166 {Billion} {Organic} {Small} {Molecules} in the
  {Chemical} {Universe} {Database} {GDB}-17.
\newblock \emph{Journal of Chemical Information and Modeling}, 52\penalty0
  (11):\penalty0 2864--2875, November 2012.
\newblock ISSN 1549-9596, 1549-960X.
\newblock \doi{10.1021/ci300415d}.
\newblock URL \url{https://pubs.acs.org/doi/10.1021/ci300415d}.

\bibitem[Ramakrishnan et~al.(2014)Ramakrishnan, Dral, Rupp, and von
  Lilienfeld]{ramakrishnan_quantum_2014}
Raghunathan Ramakrishnan, Pavlo~O. Dral, Matthias Rupp, and O.~Anatole von
  Lilienfeld.
\newblock Quantum chemistry structures and properties of 134 kilo molecules.
\newblock \emph{Scientific Data}, 1\penalty0 (1):\penalty0 140022, December
  2014.
\newblock ISSN 2052-4463.
\newblock \doi{10.1038/sdata.2014.22}.
\newblock URL \url{http://www.nature.com/articles/sdata201422}.

\bibitem[{Zhirong Wu} et~al.(2015){Zhirong Wu}, Song, Khosla, {Fisher Yu},
  {Linguang Zhang}, {Xiaoou Tang}, and Xiao]{zhirong_wu_3d_2015}
{Zhirong Wu}, Shuran Song, Aditya Khosla, {Fisher Yu}, {Linguang Zhang},
  {Xiaoou Tang}, and Jianxiong Xiao.
\newblock {3D} {ShapeNets}: {A} deep representation for volumetric shapes.
\newblock In \emph{2015 {IEEE} {Conference} on {Computer} {Vision} and
  {Pattern} {Recognition} ({CVPR})}, pages 1912--1920, Boston, MA, USA, June
  2015. IEEE.
\newblock ISBN 978-1-4673-6964-0.
\newblock \doi{10.1109/CVPR.2015.7298801}.
\newblock URL \url{https://ieeexplore.ieee.org/document/7298801/}.

\bibitem[Qi et~al.(2017{\natexlab{a}})Qi, Su, Mo, and Guibas]{qi_pointnet_2017}
Charles~R. Qi, Hao Su, Kaichun Mo, and Leonidas~J. Guibas.
\newblock Pointnet: Deep learning on point sets for 3d classification and
  segmentation.
\newblock In \emph{Proceedings of the IEEE Conference on Computer Vision and
  Pattern Recognition (CVPR)}, July 2017{\natexlab{a}}.

\bibitem[Qi et~al.(2017{\natexlab{b}})Qi, Yi, Su, and
  Guibas]{qi_pointnetplusplus_2017}
Charles~Ruizhongtai Qi, Li~Yi, Hao Su, and Leonidas~J Guibas.
\newblock Pointnet++: Deep hierarchical feature learning on point sets in a
  metric space.
\newblock In I.~Guyon, U.~Von Luxburg, S.~Bengio, H.~Wallach, R.~Fergus,
  S.~Vishwanathan, and R.~Garnett, editors, \emph{Advances in Neural
  Information Processing Systems}, volume~30. Curran Associates, Inc.,
  2017{\natexlab{b}}.
\newblock URL
  \url{https://proceedings.neurips.cc/paper_files/paper/2017/file/d8bf84be3800d12f74d8b05e9b89836f-Paper.pdf}.

\bibitem[Thompson(1994)]{thompson_angular_1994}
William~J. Thompson.
\newblock \emph{Angular momentum: an illustrated guide to rotational symmetries
  for physical systems}.
\newblock Wiley, New York, 1994.
\newblock ISBN 978-0-471-55264-2.

\bibitem[Paige and Saunders(1982{\natexlab{a}})]{paige_LSQR_1982}
Christopher~C. Paige and Michael~A. Saunders.
\newblock Lsqr: An algorithm for sparse linear equations and sparse least
  squares.
\newblock \emph{ACM Transactions on Mathematical Software}, 8\penalty0
  (1):\penalty0 43–71, Mar 1982{\natexlab{a}}.
\newblock ISSN 0098-3500.
\newblock \doi{10.1145/355984.355989}.

\bibitem[Paige and Saunders(1982{\natexlab{b}})]{paige_algorithm-583_1982}
Christopher~C. Paige and Michael~A. Saunders.
\newblock Algorithm 583: Lsqr: Sparse linear equations and least squares
  problems.
\newblock \emph{ACM Transactions on Mathematical Software}, 8\penalty0
  (2):\penalty0 195–209, Jun 1982{\natexlab{b}}.
\newblock ISSN 0098-3500.
\newblock \doi{10.1145/355993.356000}.

\bibitem[Virtanen et~al.(2020)Virtanen, Gommers, Oliphant, Haberland, Reddy,
  Cournapeau, Burovski, Peterson, Weckesser, Bright, {van der Walt}, Brett,
  Wilson, Millman, Mayorov, Nelson, Jones, Kern, Larson, Carey, Polat, Feng,
  Moore, {VanderPlas}, Laxalde, Perktold, Cimrman, Henriksen, Quintero, Harris,
  Archibald, Ribeiro, Pedregosa, {van Mulbregt}, and {SciPy 1.0
  Contributors}]{scipy}
Pauli Virtanen, Ralf Gommers, Travis~E. Oliphant, Matt Haberland, Tyler Reddy,
  David Cournapeau, Evgeni Burovski, Pearu Peterson, Warren Weckesser, Jonathan
  Bright, St{\'e}fan~J. {van der Walt}, Matthew Brett, Joshua Wilson, K.~Jarrod
  Millman, Nikolay Mayorov, Andrew R.~J. Nelson, Eric Jones, Robert Kern, Eric
  Larson, C~J Carey, {\.I}lhan Polat, Yu~Feng, Eric~W. Moore, Jake
  {VanderPlas}, Denis Laxalde, Josef Perktold, Robert Cimrman, Ian Henriksen,
  E.~A. Quintero, Charles~R. Harris, Anne~M. Archibald, Ant{\^o}nio~H. Ribeiro,
  Fabian Pedregosa, Paul {van Mulbregt}, and {SciPy 1.0 Contributors}.
\newblock {{SciPy} 1.0: Fundamental Algorithms for Scientific Computing in
  Python}.
\newblock \emph{Nature Methods}, 17:\penalty0 261--272, 2020.
\newblock \doi{10.1038/s41592-019-0686-2}.

\bibitem[Wathen(2015)]{wathen_preconditioning_2015}
A.~J. Wathen.
\newblock Preconditioning.
\newblock \emph{Acta Numerica}, 24:\penalty0 329–376, May 2015.
\newblock ISSN 0962-4929, 1474-0508.
\newblock \doi{10.1017/S0962492915000021}.

\bibitem[Benzi(2002)]{benzi_preconditioning_2002}
Michele Benzi.
\newblock Preconditioning techniques for large linear systems: A survey.
\newblock \emph{Journal of Computational Physics}, 182\penalty0 (2):\penalty0
  418–477, Nov 2002.
\newblock ISSN 0021-9991.
\newblock \doi{10.1006/jcph.2002.7176}.

\bibitem[Meng et~al.(2014)Meng, Saunders, and Mahoney]{meng_lsrn_2014}
Xiangrui Meng, Michael~A. Saunders, and Michael~W. Mahoney.
\newblock {LSRN}: {A} {Parallel} {Iterative} {Solver} for {Strongly} {Over}- or
  {Underdetermined} {Systems}.
\newblock \emph{SIAM Journal on Scientific Computing}, 36\penalty0
  (2):\penalty0 C95--C118, January 2014.
\newblock ISSN 1064-8275, 1095-7197.
\newblock \doi{10.1137/120866580}.
\newblock URL \url{http://epubs.siam.org/doi/10.1137/120866580}.

\bibitem[Drineas and Mahoney(2016)]{drineas_randnla_2016}
Petros Drineas and Michael~W. Mahoney.
\newblock Randnla: randomized numerical linear algebra.
\newblock \emph{Communications of the ACM}, 59\penalty0 (6):\penalty0 80–90,
  May 2016.
\newblock ISSN 0001-0782.
\newblock \doi{10.1145/2842602}.

\bibitem[Chowdhury et~al.(2018)Chowdhury, Yang, and
  Drineas]{chowdhury_iterative_2018}
Agniva Chowdhury, Jiasen Yang, and Petros Drineas.
\newblock An iterative, sketching-based framework for ridge regression.
\newblock In \emph{Proceedings of the 35th International Conference on Machine
  Learning}, page 989–998. PMLR, Jul 2018.
\newblock URL \url{https://proceedings.mlr.press/v80/chowdhury18a.html}.

\bibitem[Kacham and Woodruff(2022)]{kacham_sketching_2022}
Praneeth Kacham and David Woodruff.
\newblock Sketching algorithms and lower bounds for ridge regression.
\newblock In \emph{Proceedings of the 39th International Conference on Machine
  Learning}, page 10539–10556. PMLR, Jun 2022.
\newblock URL \url{https://proceedings.mlr.press/v162/kacham22a.html}.

\end{thebibliography}
